%% file: main.tex
\documentclass[10pt]{article} 
\usepackage[preprint]{tmlr}

\input{math_commands.tex}

\usepackage{hyperref}
\usepackage{url}
\usepackage{graphicx}
\usepackage{array}
\usepackage{booktabs}
\usepackage{placeins}
\usepackage{wrapfig}

\title{Pre-Intervention Prediction of Sparse Autoencoder Steering Side Effects}

\author{\name Evan Duan \email evanduan@umich.edu \\
}

\def\month{MM}  
\def\year{YYYY} 
\def\openreview{\url{https://openreview.net/forum?id=XXXX}} 

\begin{document}
\raggedbottom

\maketitle

\begin{abstract}

Sparse autoencoder (SAE) features are increasingly used to steer language models, but feature steering is rarely clean: the same intervention can behave inconsistently across contexts and perturb unrelated features. We introduce a pre-intervention screening framework for forecasting SAE steering side effects from feature statistics computed before steering. We operationalize side effects along two axes of steering modularity, effect stability and collateral spread, and evaluate GPT-2-small, Pythia-70M-deduped, Gemma-2-2B, and Llama-3.1-8B across ReLU, JumpReLU, and TopK SAE dictionaries. Across these settings, decoder geometry, activation statistics, co-activation structure, and direct-logit footprint predict steering modularity better than frequency-only and activation-magnitude baselines. The signal is strongest in GPT-2-small, Pythia-70M, and Llama-3.1-8B, where it survives residualization against magnitude-related confounds, and weaker in Gemma-2-2B. Held-out screening shows that ranking unseen features by predicted cleanliness can select features that steer more cleanly on fresh contexts, but the successful axis varies by setting: GPT-2 improves most cleanly, Pythia improves mainly on stability, Llama mainly on collateral, and Gemma only partially. A controlled Llama Scope width comparison shows that the predictive signal persists under a 32K-to-128K dictionary-width change, although the screening payoff becomes less stable. Overall, SAE steering side effects are predictable in advance, but the useful predictor signature and transferred modularity axis are model- and dictionary-setting dependent.

\end{abstract}

\input{methodology.tex}

\bibliographystyle{tmlr}
\bibliography{main}

\clearpage
\appendix
\setcounter{table}{0}
\renewcommand{\thetable}{A\arabic{table}}
\renewcommand{\theHtable}{appendixtable.\arabic{table}}
\renewcommand{\tablename}{Appendix Table}
\input{appendix_table_a1.tex}
\FloatBarrier

\setlength{\textfloatsep}{6pt plus 1pt}
\setlength{\floatsep}{6pt plus 1pt}
\setlength{\intextsep}{6pt plus 1pt}
\setcounter{table}{0}
\renewcommand{\thetable}{B\arabic{table}}
\renewcommand{\theHtable}{appendixtableB.\arabic{table}}
\input{appendix_predictive_evaluation.tex}
\FloatBarrier

\setlength{\textfloatsep}{12pt plus 2pt}
\setlength{\floatsep}{12pt plus 2pt}
\setlength{\intextsep}{12pt plus 2pt}
\setlength{\abovecaptionskip}{6pt plus 1pt}
\setlength{\belowcaptionskip}{4pt plus 1pt}

\clearpage
\setcounter{table}{0}
\renewcommand{\thetable}{C\arabic{table}}
\renewcommand{\theHtable}{appendixtableC.\arabic{table}}
\setcounter{figure}{0}
\renewcommand{\thefigure}{C\arabic{figure}}
\renewcommand{\theHfigure}{appendixfigureC.\arabic{figure}}
\input{appendix_predictor_label_diagnostics.tex}

\FloatBarrier

\clearpage
\setcounter{table}{0}
\renewcommand{\thetable}{D\arabic{table}}
\renewcommand{\theHtable}{appendixtableD.\arabic{table}}
\setcounter{figure}{0}
\renewcommand{\thefigure}{D\arabic{figure}}
\renewcommand{\theHfigure}{appendixfigureD.\arabic{figure}}
\makeatletter
\setlength{\@fptop}{0pt}
\setlength{\@fpsep}{8pt plus 1fil}
\setlength{\@fpbot}{0pt plus 1fil}
\makeatother
\input{appendix_heldout_screening_details.tex}

\FloatBarrier
\clearpage
\input{appendix_additional_metric_predictor_definitions.tex}

\FloatBarrier
\clearpage
\setcounter{table}{0}
\renewcommand{\thetable}{F\arabic{table}}
\renewcommand{\theHtable}{appendixtableF.\arabic{table}}
\input{appendix_llama_replication.tex}

\FloatBarrier
\setcounter{table}{0}
\renewcommand{\thetable}{G\arabic{table}}
\renewcommand{\theHtable}{appendixtableG.\arabic{table}}
\input{appendix_llama_width_control.tex}

\FloatBarrier
\input{appendix_reproducibility_hardware.tex}

\end{document}

%% file: math_commands.tex
\usepackage{amsmath,amsfonts,bm}

\def\eqref#1{equation~\ref{#1}}

\def\1{\bm{1}}

\DeclareMathAlphabet{\mathsfit}{\encodingdefault}{\sfdefault}{m}{sl}
\SetMathAlphabet{\mathsfit}{bold}{\encodingdefault}{\sfdefault}{bx}{n}



%% file: methodology.tex
\section{Introduction}

Sparse autoencoders (SAEs) have become a central tool for decomposing language-model activations into sparse, more interpretable features \citep{bricken2023monosemanticity,cunningham2023sparse,templeton2024scaling}. A natural application of these features is steering: adding a feature's decoder direction to the residual stream to push the model's behavior along an interpretable axis \citep{turner2023steering,zou2023representation}. The appeal is targeted control: one feature, one direction, and an effect that can be interpreted.

In practice, however, steering a feature is rarely clean. The same intervention can push the model's outputs in inconsistent directions across contexts, and it can disturb features or behaviors unrelated to the intended target. These side effects matter for reliability- and safety-oriented uses of steering, where an edit is useful only if it is predictable and well-contained. Yet side effects are usually discovered only after intervention, by running the steering operation across many inputs and measuring what changed. There is no cheap, a priori guide to which features will steer cleanly, and whether such side effects are predictable in advance has not been systematically established. A reliable pre-intervention screen would let practitioners select better-behaved features without an expensive per-feature intervention sweep.

We ask whether the side effects of SAE feature steering can be forecast before intervention from cheap statistics of the feature itself. We operationalize side effects along two measurable axes of steering modularity: stability, the cross-context consistency of the steering effect's direction, and collateral spread, the breadth and magnitude of change induced in other features. Our hypothesis is that intervention-free properties of a feature, including decoder geometry, activation distribution, co-activation structure, and direct-logit footprint, already contain information about how modular its steering effect will be.

To test this, we measure steering labels for 300 SAE features in each of four model/SAE settings: GPT-2-small \citep{radford2019language}, Pythia-70M-deduped \citep{biderman2023pythia}, Gemma-2-2B \citep{team2024gemma}, and Llama-3.1-8B \citep{grattafiori2024llama}. These settings span multiple model architectures and dictionary families: standard ReLU SAEs \citep{bricken2023monosemanticity,cunningham2023sparse}, JumpReLU Gemma Scope SAEs \citep{lieberum2024gemma,rajamanoharan2024jumping}, and TopK Llama Scope residual-stream SAEs \citep{he2024llama}. For each feature, we apply a sign-stable additive intervention across a mixed set of contexts, record stability and collateral labels, and compute intervention-free predictors grouped into decoder-geometry, activation, co-activation, and direct-logit families. We then evaluate two questions. The predictive evaluation asks whether cross-validated regression on cheap predictors recovers measured steering labels and improves over frequency-only and activation-magnitude-only baselines. The held-out screening evaluation asks whether ranking previously unseen features by predicted cleanliness selects features that steer more stably and with less collateral on fresh contexts. All models and SAEs are pretrained; we train no SAEs.

Our findings are threefold, and we report them at a deliberately calibrated strength. First, cheap intervention-free predictors forecast steering modularity: a predictor set that excludes activation magnitude improves cross-validated rank correlation over frequency-only and activation-magnitude-only baselines. This relationship is strongest and most robust in GPT-2-small, Pythia-70M, and Llama-3.1-8B, and weaker and more magnitude-entangled in Gemma-2-2B. Second, the dominant predictor is not the same across models: decoder crowding leads in GPT-2-small, the direct-logit footprint leads for collateral in Pythia-70M, broad encoder geometry leads in Gemma-2-2B, and Llama-3.1-8B shows strong geometry and direct-logit signal. This separates method-level transfer from mechanism-level transfer: the general screening procedure transfers more reliably than any single predictor signature. Third, held-out screening improves steering cleanliness on fresh contexts, but the successful axis varies by model. In GPT-2-small, predicted-clean features are both more stable and lower in collateral; in Pythia-70M, they are reliably more stable while the collateral benefit is not statistically confirmed; in Gemma-2-2B, the benefit is partial and limited to collateral; and in Llama-3.1-8B, the benefit appears on the collateral axis rather than the stability axis.

Taken together, these results indicate that SAE steering side effects are predictable in advance, but that the predictive signature and transferred modularity axis are model- and dictionary-setting dependent rather than universal. We therefore frame our contribution as a method for cheap, pre-intervention screening of features for steering modularity, while treating the dominant predictor in any given model as an empirical, model-specific finding rather than a general mechanism. We make the following contributions:
\begin{itemize}
\item We introduce a framework for predicting SAE steering side effects before intervention, operationalizing side effects as two measurable axes---stability and collateral spread---and forecasting them from intervention-free feature statistics.
\item We present a cross-model predictive evaluation over four model/SAE settings, showing that cheap predictors excluding activation magnitude beat frequency and activation-magnitude baselines and survive confound residualization---strongly in GPT-2-small, Pythia-70M, and Llama-3.1-8B, and weakly in Gemma-2-2B.
\item We provide a held-out screening evaluation demonstrating that ranking unseen features by predicted cleanliness selects features that steer more cleanly on fresh contexts---on both axes in GPT-2-small, on stability in Pythia-70M, on collateral in Llama-3.1-8B, and partially in Gemma-2-2B.
\item We draw an explicit distinction between method-level and mechanism-level transfer, with evidence that the dominant predictor signature and successful screening axis are model- and dictionary-setting dependent rather than universal.
\end{itemize}

\section{Related Work}
\label{sec:related}

Sparse autoencoders (SAEs) have become a leading method for decomposing language-model activations into sparse, approximately monosemantic features \citep{bricken2023monosemanticity,cunningham2023sparse}. This approach is motivated by \emph{superposition}, the hypothesis that models represent more features than they have dimensions by placing them in overlapping linear directions \citep{elhage2022toy}, and sits within a broader mechanistic-interpretability program that aims to reverse-engineer model computation into human-understandable components \citep{elhage2021framework}. Subsequent work scaled SAEs to production-scale models and showed that recovered features can be interpretable and causally relevant \citep{templeton2024scaling}, with automated pipelines used to label features at scale \citep{bills2023language}. Our work takes pretrained SAEs as given: we train no SAEs, and instead study the downstream behavior of their features under intervention.

Recent SAE infrastructure has made it possible to study features beyond a single model. Gemma Scope released a comprehensive suite of JumpReLU SAEs across Gemma 2 \citep{lieberum2024gemma}, while JumpReLU SAEs improve the sparsity--fidelity tradeoff over standard ReLU SAEs \citep{rajamanoharan2024jumping}. Parallel work has also scaled and standardized SAE training and evaluation \citep{gao2025scaling}. This infrastructure allows us to ask the same predictive question across multiple model/SAE settings spanning ReLU, JumpReLU, and TopK dictionaries. Because model and SAE family are not fully crossed in our design, we use these comparisons to study transfer across dictionary settings rather than to causally isolate SAE-family effects.

Recent work has also studied predictable structure in SAE reconstruction error. For example, \citet{engels2024decomposing} analyze the ``dark matter'' of SAEs: the component of activations not captured by sparse reconstruction, and show that substantial parts of SAE error can be predicted from the original activation. Our work studies a different prediction target. Rather than predicting SAE 
reconstruction error at the activation level, we predict the feature-level side effects of steering, including cross-context stability and downstream collateral spread.

Feature directions are also widely used to steer model behavior. Activation addition biases the residual stream with a direction that shifts outputs along an interpretable axis without optimization \citep{turner2023steering}; contrastive activation addition derives such directions from differences between paired positive and negative examples \citep{panickssery2023steering}; and representation engineering frames the broader program of reading and manipulating population-level representations \citep{zou2023representation}. SAE features offer a particularly direct handle for this kind of control, since each feature has a decoder direction that can be added to the stream \citep{templeton2024scaling}. Recent work further suggests that SAE steering quality depends strongly on selecting appropriate features, motivating feature-selection methods for steering \citep{arad2025saes}. The intervention we study, additive steering of a single SAE feature, is an instance of this paradigm, but our question concerns its reliability rather than its mechanism of action.

A recurring observation across the steering and model-editing literature is that interventions are not uniformly reliable. Steering vectors can fail to generalize across prompts and behaviors, and can perturb outputs in unintended ways \citep{turner2023steering,panickssery2023steering}. SAE-based steering has also been shown to improve targeted safety behaviors while degrading unrelated capabilities, directly motivating collateral-effect evaluation \citep{o2024steering}. Related control paradigms such as direct weight editing also face analogous ripple effects on knowledge or behaviors the edit was not meant to affect \citep{meng2022locating}. These reliability properties are usually characterized \emph{after} intervention, by steering or editing and then measuring the effects across inputs. To our knowledge, prior work does not provide an a priori, intervention-free way to predict which SAE features will steer stably and with limited collateral, nor does it study how such predictability changes across model/SAE settings.

We address this gap by treating steering side effects as a prediction target. Rather than proposing a new steering method or a new SAE, we ask whether cheap, intervention-free statistics of a feature, including decoder geometry, activation distribution, co-activation structure, and direct-logit footprint, forecast its steering stability and collateral spread before intervention. We further distinguish \emph{method-level transfer}, where the general procedure of predicting and screening on these statistics carries across models, from \emph{mechanism-level transfer}, where a single predictor signature is universal. This situates our contribution alongside, but distinct from, work that builds SAEs, work that uses feature directions to steer models, and work that documents steering limitations post hoc.

\section{Pre-Intervention Screening Framework}

\begin{figure}[!htbp]
\begin{center}
\includegraphics[width=0.95\linewidth]{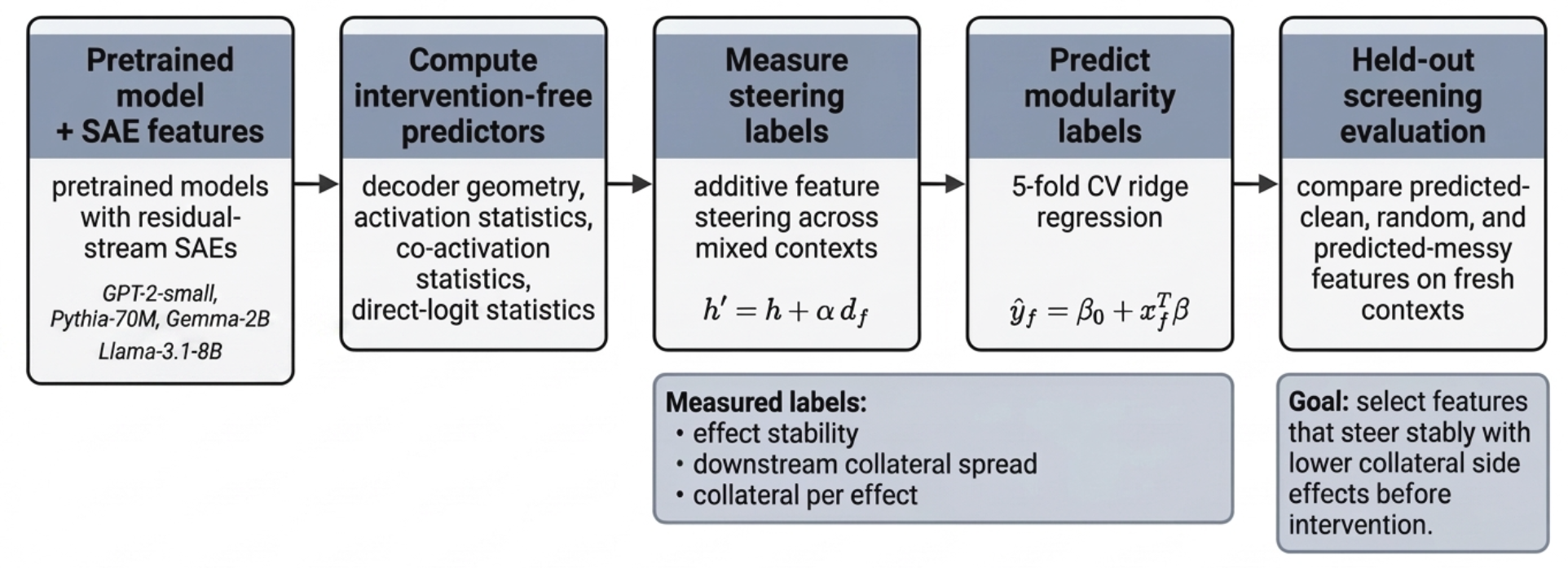}
\end{center}
\caption{\textbf{Method overview.} SAE predictors are used to forecast steering stability and collateral effects before intervention.}
\label{fig:pipeline}
\end{figure}

We evaluate whether SAE steering side effects can be predicted before intervention. Figure~\ref{fig:pipeline} is the summary of the pipeline. We use four model/SAE settings, summarized in Table~\ref{tab:main-config}. Because model architecture and dictionary family are not fully crossed, we treat them as model/SAE settings rather than independent model--SAE-family factors. All models and SAEs are pretrained, with no fine-tuning or SAE training.

\paragraph{Controlled dictionary-width check.}
In addition to the four primary settings, we run a controlled Llama Scope width comparison to probe dictionary dependence. This check holds the base model, hook sites, dataset, feature count, context count, intervention, and evaluation protocol fixed, while replacing the 32K Llama Scope residual-stream SAEs (\texttt{l16r\_8x}, \texttt{l20r\_8x}) with 128K SAEs (\texttt{l16r\_32x}, \texttt{l20r\_32x}). We use this comparison as a robustness check for dictionary granularity, not as a full causal isolation of SAE-family effects.

\begin{table}[t]
\caption{Main experimental configuration.}
\label{tab:main-config}
\begin{center}
\small
\begin{tabular}{@{}p{0.22\linewidth}p{0.70\linewidth}@{}}
\multicolumn{1}{c}{\bf Model} & \multicolumn{1}{c}{\bf Configuration} \\
\hline \\
GPT-2-small & Primary: \texttt{blocks.8.hook\_resid\_pre}; downstream: \texttt{blocks.10.hook\_resid\_pre}; 300 features; 2{,}048 contexts; length 48. \\
Pythia-70M-\allowbreak deduped & Primary: \texttt{blocks.4.hook\_resid\_post}; downstream: \texttt{blocks.5.hook\_resid\_post}; 300 features; 2{,}048 contexts; length 48. \\
Gemma-2-2B & Primary: Gemma Scope layer-12 residual SAE; downstream: Gemma Scope layer-16 residual SAE; 300 features; 2{,}048 contexts; length 48. \\
Llama-3.1-8B & Primary: \texttt{blocks.16.hook\_resid\_post} / Llama Scope \texttt{l16r\_8x}; downstream: \texttt{blocks.20.hook\_resid\_post} / Llama Scope \texttt{l20r\_8x}; 300 features; 2{,}048 contexts; length 48. \\
\end{tabular}
\end{center}
\end{table}

\subsection{Sparse autoencoder features}

Let $M_\theta$ be a transformer language model and let $h_{\ell,t}(x) \in \mathbb{R}^{d_{\mathrm{model}}}$ denote the activation at hook site $\ell$, token position $t$, and input context $x$. A pretrained sparse autoencoder maps this activation to a feature vector
\[
z_{\ell,t}(x) = E(h_{\ell,t}(x)) \in \mathbb{R}^{m},
\]
with decoder directions $d_f \in \mathbb{R}^{d_{\mathrm{model}}}$ for features $f \in \{1,\ldots,m\}$. We use pretrained SAEs only; no SAE is trained in this work.

For each model, we select a primary SAE site where steering is applied and, when available, a later downstream SAE site where collateral feature movement is measured. Exact SAE releases and hook sites are reported in Appendix~A.

\subsection{Corpus, feature sampling, and context selection}

We use Wikitext-103 for predictor construction and steering-label construction. Texts are deduplicated after whitespace normalization and tokenized into fixed-length contexts. For each model, we construct 2{,}048 contexts and select 300 SAE features after filtering out degenerate features with near-zero or near-universal firing.

For feature $f$, let $a_f(x_i)$ denote its activation at the final token position of context $x_i$. Its firing frequency is
\[
\mathrm{freq}(f) =
\frac{1}{N}
\sum_{i=1}^{N}
\mathbf{1}[a_f(x_i) > \epsilon_{\mathrm{fire}}],
\]
where $\epsilon_{\mathrm{fire}} = 10^{-6}$. Features are sampled from the non-degenerate range of the firing-frequency distribution.

For each selected feature, steering labels are measured on mixed context sets. These include top-activating contexts, random contexts, and low- or non-firing contexts. We use the mixed set as the primary label context set so that stability is measured across both natural and less natural activation regimes.

\subsection{Additive feature steering}

For feature $f$ with decoder direction $d_f$, we apply additive steering at the final token position:
\[
h'_{\ell,t}(x;f,\alpha)=h_{\ell,t}(x)+\alpha d_f,\qquad \alpha=1.0.
\]
This avoids clamp-style sign flips, where an intervention may suppress high-activation contexts but amplify low-activation contexts. For each context $x_i$, we compute clean and steered logits and downstream SAE activations, defining
\[
\Delta \ell_i^{(f)}=\ell_i^{(f)}-\ell_i,\qquad
\Delta u_i^{(f)}=u_i^{(f)}-u_i.
\]
Here $\Delta \ell_i^{(f)}$ is the logit effect and $\Delta u_i^{(f)}$ is the downstream SAE feature effect. Downstream collateral metrics are computed on a fixed panel of frequently active downstream features.

\subsection{Steering labels}

We measure steering modularity along two axes: effect stability and collateral spread. Let $\mathcal{C}_f$ denote the mixed context set for feature $f$, and let
\[
\overline{\Delta \ell}^{(f)}
=
|\mathcal{C}_f|^{-1}
\sum_{x_i \in \mathcal{C}_f}
\Delta \ell_i^{(f)}
\]
be the mean logit effect. We define signed and absolute stability as
\[
S_f^{\mathrm{signed}}
=
|\mathcal{C}_f|^{-1}
\sum_{x_i \in \mathcal{C}_f}
\cos\!\left(
\Delta \ell_i^{(f)},
\overline{\Delta \ell}^{(f)}
\right),
\qquad
S_f^{\mathrm{abs}}
=
|\mathcal{C}_f|^{-1}
\sum_{x_i \in \mathcal{C}_f}
\left|
\cos\!\left(
\Delta \ell_i^{(f)},
\overline{\Delta \ell}^{(f)}
\right)
\right|.
\]
The signed score penalizes opposite-direction effects, while the absolute score measures whether effects lie on a consistent one-dimensional axis. The held-out screening evaluation uses $S_f^{\mathrm{abs}}$ as the primary stability metric.

Collateral spread is measured in a downstream SAE feature panel $\mathcal{P}$. For threshold $\tau$, we define downstream collateral count, mean logit-effect magnitude, and collateral per effect as
\[
C_{f,\tau}
=
|\mathcal{C}_f|^{-1}
\sum_{x_i \in \mathcal{C}_f}
\sum_{j \in \mathcal{P}}
\mathbf{1}\!\left[
|\Delta u_{i,j}^{(f)}| > \tau
\right],
\qquad
E_f
=
|\mathcal{C}_f|^{-1}
\sum_{x_i \in \mathcal{C}_f}
\|\Delta \ell_i^{(f)}\|_2,
\qquad
\widetilde{C}_{f,\tau}
=
\frac{C_{f,\tau}}{E_f+\varepsilon}.
\]
The primary collateral metric is $\widetilde{C}_{f,0.05}$. Additional label definitions, including KL shift and auxiliary effect-magnitude metrics, are provided in Appendix~E.

\subsection{Intervention-free predictors}

We compute all predictors before steering and group them into decoder-geometry, activation, co-activation, and direct-logit statistics. We evaluate seven predictor sets: frequency-only, activation-magnitude-only, geometry-only, direct-logit-only, co-activation-only, full no-magnitude, and full-all. The main comparison asks whether full no-magnitude improves over frequency-only and activation-magnitude baselines. Detailed definitions are provided in Appendix \ref{app:additional-metric-predictor-definitions}.

\subsection{Mechanistic motivation for the predictor families}

The predictor families are motivated by a local-linear view of steering. Let $\phi_j(h)$ be the downstream activation of feature $j$ as a function of the residual-stream state $h$. Steering feature $f$ adds its decoder direction $d_f$:
\[
h' = h + \alpha d_f .
\]
For small local perturbations, the downstream change in feature $j$ can be approximated by a first-order expansion,
\[
\Delta u_j^{(f)}
=
\phi_j(h+\alpha d_f)-\phi_j(h)
\approx
\alpha \nabla_h \phi_j(h)^\top d_f .
\]
Thus, collateral spread is expected to be larger when $d_f$ has nontrivial projection onto many downstream sensitivity directions:
\[
C_{f,\tau}
\approx
\sum_{j \in \mathcal{P}}
\mathbf{1}\!\left[
\left|
\alpha \nabla_h \phi_j(h)^\top d_f
\right| > \tau
\right].
\]
This motivates decoder-geometry and co-activation predictors: crowded decoder neighborhoods or broad co-activation patterns make steering more likely to overlap with other feature directions, increasing collateral movement. Applying the same local-linear view to logits,
\[
\Delta \ell^{(f)} \approx \alpha J_\ell(h)d_f,
\]
motivates direct-logit predictors and stability metrics: broad logit footprints suggest diffuse output perturbations, while stability depends on whether $J_\ell(h)d_f$ remains directionally consistent across contexts. These approximations are motivational only; the experiments test which signals actually forecast steering modularity in each model/SAE setting.

\subsection{Predictive evaluation}

For feature $f$, let $x_f \in \mathbb{R}^p$ be its predictor vector and $y_f$ its measured steering label. For each predictor set, we fit ridge regression with standardized predictors and evaluate five-fold cross-validated Spearman correlation:
\[
\hat{y}_f=\beta_0+x_f^\top\beta,\qquad
\rho=\operatorname{Spearman}\!\left(
\{\hat{y}_f\}_{f\in\mathcal{H}},
\{y_f\}_{f\in\mathcal{H}}
\right),
\]
where $\mathcal{H}$ is the held-out fold. To summarize gains over baselines, we report
\[
\Delta_{\mathrm{freq}}
=
\rho_{\mathrm{no\text{-}mag}}-\rho_{\mathrm{freq}},
\qquad
\Delta_{\mathrm{actmag}}
=
\rho_{\mathrm{no\text{-}mag}}-\rho_{\mathrm{actmag}}.
\]
These quantities test whether predictors that exclude activation magnitude outperform the frequency-only and activation-magnitude baselines.

\subsection{Residualized robustness evaluation}

To test whether predictors merely recover effect magnitude or natural activation strength, we residualize stability labels against nuisance variables. For a stability label $y_f$, we fit
\[
y_f
=
\gamma_0
+
\gamma_1 E_f
+
\gamma_2 c_f
+
\gamma_3 \bar{a}_f
+
\eta_f,
\]
where $E_f$ is effect magnitude, $c_f$ is the intervention value, and $\bar{a}_f$ is the feature's natural activation level. We then use the residual $y_f^{\mathrm{resid}} = \eta_f$ as the prediction target and repeat the same cross-validated ridge evaluation.

\subsection{Held-out screening evaluation}

The held-out screening evaluation tests whether predictor scores can serve as a practical feature-selection rule. For each model and selector, we split the measured feature set into a training set and a held-out selection pool. A ridge model is trained to predict a clean-steering score,
\[
Q_f
=
Z(S_f^{\mathrm{abs}})
-
Z\!\left(
\log(1 + \widetilde{C}_{f,0.05})
\right),
\]
where $Z(\cdot)$ denotes z-scoring across features. Higher $Q_f$ corresponds to higher stability and lower collateral per unit effect.

Features in the held-out pool are ranked by predicted $\hat{Q}_f$. We assign the top-ranked features to the predicted-clean group, the bottom-ranked features to the predicted-messy group, and sample random controls from the remaining pool, after filtering to a central predicted-effect range to match groups on expected effect magnitude.

Selected features are evaluated on fresh Wikitext validation contexts not used for steering-label construction. We test predicted-clean versus predicted-messy groups with two-sided Mann--Whitney tests on realized stability and collateral per effect. A result is labeled strong if predicted-clean is significantly better on both axes, partial if it is better on one axis, and weak otherwise.

\section{Results}

This section reports the empirical findings for four model/SAE settings---GPT-2-small, Pythia-70M-deduped, Gemma-2-2B, and Llama-3.1-8B---across two evaluation settings: a predictive evaluation over measured steering labels and a held-out screening evaluation on newly selected features and contexts. All reported correlations are Spearman rank correlations. Predictive-model performance is reported as five-fold cross-validated (CV) Spearman correlation between predicted and measured modularity over 300 features per model. Modularity is measured along two axes: stability, the cross-context consistency of the steering-effect direction, and collateral spread, the breadth and magnitude of induced change in a downstream feature panel.

We summarize each result using a rule fixed in advance. For the screening evaluation, a contrast is labeled strong if the predicted-clean group differs significantly from the predicted-messy group (two-sided Mann--Whitney, $p < 0.05$) on both pre-specified axes---stability and collateral per effect---partial if on exactly one axis, and weak if on neither. For the predictive evaluation we describe results by the magnitude of the no-magnitude model's improvement over the baselines rather than by a categorical label.

\subsection{Predictive performance across models}

\begin{figure}[!htbp]
\centering
\includegraphics[width=\textwidth]{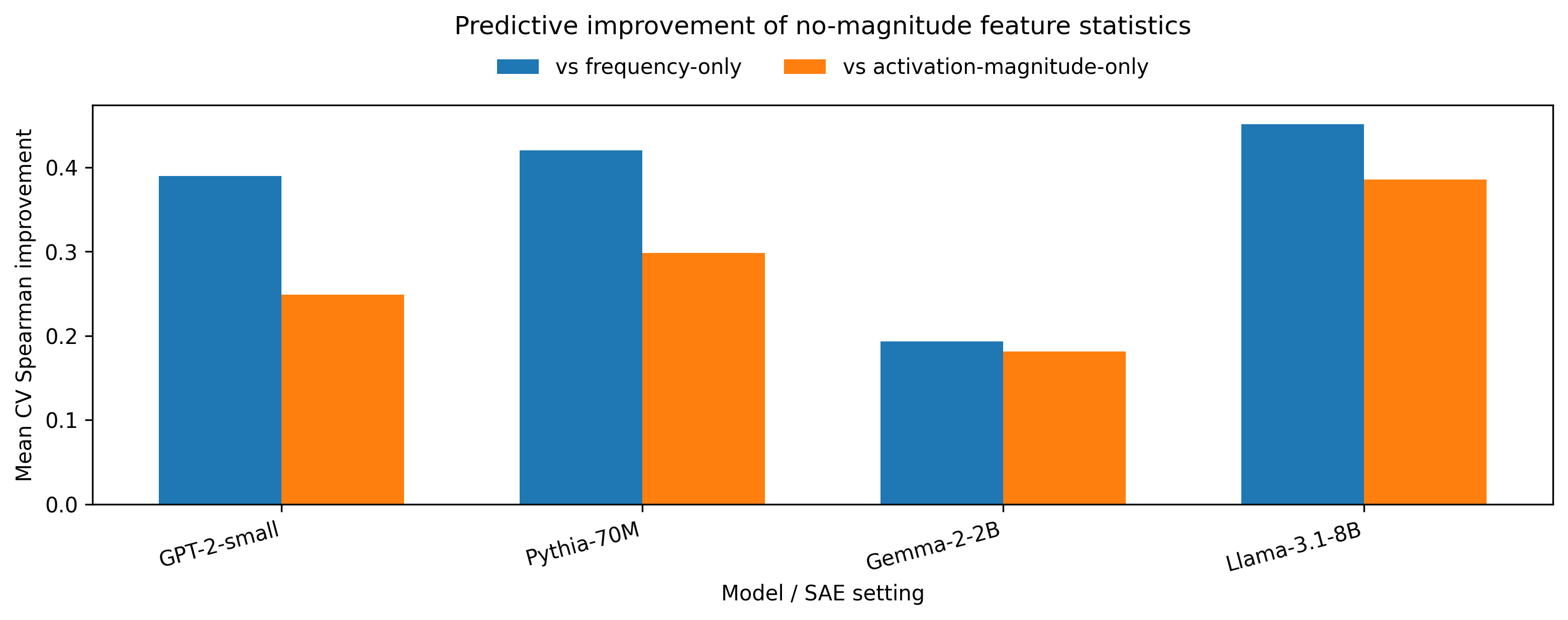}
\caption{Mean CV Spearman gain over frequency and activation-magnitude baselines}
\label{fig:predictive-performance}
\end{figure}

In all four model/SAE settings, the no-magnitude predictor set exceeded both the frequency-only and activation-magnitude-only baselines (Figure~\ref{fig:predictive-performance}). Averaged over the eight primary targets, the no-magnitude model improved CV Spearman over the frequency baseline by $+0.390$ in GPT-2-small, $+0.421$ in Pythia-70M, $+0.193$ in Gemma-2-2B, and $+0.451$ in Llama-3.1-8B, and over the activation-magnitude baseline by $+0.249$, $+0.299$, $+0.182$, and $+0.386$, respectively. Thus, the strongest baseline improvements occur in Llama and Pythia, while Gemma shows the weakest predictive signal. The geometry-only set also improved over the frequency baseline in all settings, but by a smaller margin than the full no-magnitude set.

On signed stability, geometry-only prediction reached CV Spearman of $0.517$ (GPT-2), $0.389$ (Pythia), $0.293$ (Gemma), and $0.368$ (Llama), while the no-magnitude set reached $0.554$, $0.590$, $0.396$, and $0.447$. On the downstream-count collateral target, geometry-only reached $0.512$ (GPT-2), $0.087$ (Pythia), $0.329$ (Gemma), and $0.540$ (Llama), while the no-magnitude set reached $0.556$, $0.613$, $0.381$, and $0.631$. Llama showed a strong predictive signal, especially on collateral targets, where both geometry and direct-logit predictors were informative. In Gemma, by contrast, the absolute-cosine stability target was not predictable by any set (no-magnitude CV Spearman $0.028$, with the frequency, activation-magnitude, geometry, and crowding sets all near zero or negative), whereas the signed-stability and downstream-count targets retained measurable signal. Across the primary targets, GPT-2, Pythia, and Llama showed substantially larger improvements over baseline than Gemma. The complete per-target comparison across all predictor sets is given in Appendix~\ref{tab:appendix-baseline-comparison}.

\subsection{Dominant predictor by model}

\begin{figure}[!htbp]
\centering
\includegraphics[width=0.92\linewidth]{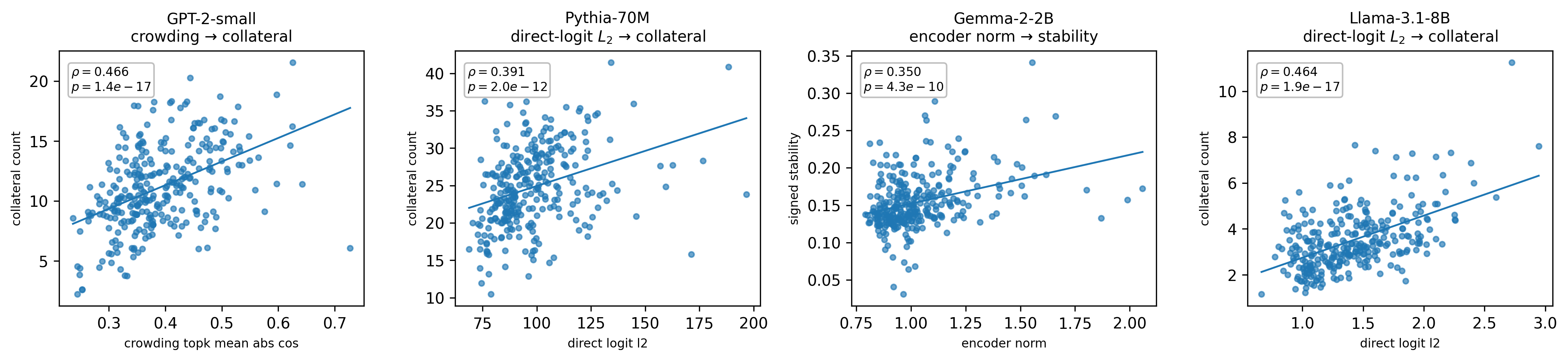}
\caption{Representative univariate predictor--label relationships across the four model/SAE settings. Scatter plots show diagnostic primary-target relationships for GPT-2-small, Pythia-70M, Gemma-2-2B, and Llama-3.1-8B, illustrating that the dominant predictor signature differs across model and dictionary settings. Spearman correlations and fitted linear trends are shown for visualization.}
\label{fig:representative-univariate-relationships.}
\end{figure}

The strongest univariate predictor differed across model/SAE settings (Table~\ref{tab:dominant-predictor-by-model}; Figure~\ref{fig:representative-univariate-relationships.}). In GPT-2, decoder crowding led for both stability and collateral, reaching $\rho = 0.466$ on downstream collateral ($p = 1.4 \times 10^{-17}$). In Pythia, direct-logit $L_2$ led for collateral ($\rho = 0.391$, $p = 2.0 \times 10^{-12}$), while decoder crowding led for stability ($\rho = -0.360$, $p = 1.3 \times 10^{-10}$); geometry-only prediction of collateral was weak (CV Spearman $0.087$). In Gemma, encoder norm led for signed stability ($\rho = 0.350$, $p = 4.3 \times 10^{-10}$). In Llama, decoder norm mainly tracked intervention magnitude ($\rho = 0.727$, $p = 1.1 \times 10^{-50}$), while side-effect-relevant targets were led by direct-logit $L_2$ for downstream collateral ($\rho = 0.464$, $p = 1.9 \times 10^{-17}$) and decoder norm for signed stability ($\rho = 0.366$, $p = 6.0 \times 10^{-11}$). Full univariate correlations appear in Appendix Table~\ref{tab:appendix-univariate-correlations}.

\begin{table}[t]
\caption{Strongest single pre-intervention predictor per model (Spearman correlation over 300 features). Pythia is reported for both label families because the leading predictor differs between them.}
\label{tab:dominant-predictor-by-model}
\begin{center}
\small
\setlength{\tabcolsep}{3pt}
\begin{tabular}{@{}>{\raggedright\arraybackslash}p{0.16\linewidth}>{\raggedright\arraybackslash}p{0.33\linewidth}>{\raggedright\arraybackslash}p{0.25\linewidth}rr@{}}
\hline
\textbf{Model} & \textbf{Predictor} & \textbf{Target} & \textbf{$\rho$} & \textbf{$p$}\\
\hline
GPT-2-small & decoder crowding (top-$k$ mean abs. cos) & downstream count (collateral) & $0.466$ & $1.4 \times 10^{-17}$\\
Pythia-70M & direct-logit footprint ($L_2$) & downstream count (collateral) & $0.391$ & $2.0 \times 10^{-12}$\\
Pythia-70M & decoder crowding (top-$k$ mean abs. cos) & stability & $-0.360$ & $1.3 \times 10^{-10}$\\
Gemma-2-2B & encoder norm & stability (signed) & $0.350$ & $4.3 \times 10^{-10}$\\
Llama-3.1-8B & direct-logit footprint ($L_2$) & downstream count (collateral) & $0.464$ & $1.9 \times 10^{-17}$\\
Llama-3.1-8B & decoder norm & stability (signed) & $0.366$ & $6.0 \times 10^{-11}$\\
\hline
\end{tabular}
\end{center}
\end{table}

\subsection{Robustness to confound residualization}
\label{sec:residual}

To test whether predictors merely track intervention magnitude, we residualized stability labels against effect $L_2$, intervention value, and natural activation, then re-evaluated prediction on the residuals. The signal was retained in GPT-2, Pythia, and Llama, but not Gemma. In Llama, the no-magnitude set reached CV Spearman $0.366$ on residualized signed stability and $0.290$ on residualized absolute stability, while frequency-only and activation-magnitude baselines were negative. Full residualized results are given in Appendix~\ref{tab:appendix-residualized-results}. As an additional dictionary-width check, we repeated the Llama experiment with 128K Llama Scope SAEs while holding the model, hook sites, data, feature count, context count, and intervention fixed. Predictive gains persisted under this 4$\times$ width increase ($+0.420$ over frequency, $+0.266$ over activation magnitude), though attenuated relative to 32K (Appendix~\ref{app:llama-width-control}).

\FloatBarrier

\begin{table}[!h]
\caption{Held-out screening performance by model and selector, evaluated on fresh contexts. Stability is mean absolute cosine to the average effect (higher is cleaner); collateral is downstream count per unit effect (lower is cleaner). The $p$-columns report two-sided Mann--Whitney tests for the predicted-clean vs. predicted-messy contrast on each axis. ``n.s.'' denotes $p \geq 0.05$.}
\label{tab:heldout-screening-performance}
\begin{center}
\scriptsize
\setlength{\tabcolsep}{2pt}
\resizebox{\linewidth}{!}{%
\begin{tabular}{@{}llrrrrrrrrl@{}}
\hline
\textbf{Model} & \textbf{Selector} & \textbf{$n$/group} & \textbf{Contexts} & \textbf{Clean stability} & \textbf{Messy stability} & \textbf{$p$ stability} & \textbf{Clean coll./effect} & \textbf{Messy coll./effect} & \textbf{$p$ collateral} & \textbf{Verdict}\\
\hline
GPT-2-small & \texttt{geometry\_only} & 32 & 512 & $0.610$ & $0.538$ & $3.98 \times 10^{-6}$ & $3.486$ & $4.010$ & $0.0198$ & strong\\
Pythia-70M & \texttt{full\_no\_magnitude} & 30 & 512 & $0.879$ & $0.803$ & $1.00 \times 10^{-8}$ & $0.682$ & $0.707$ & $0.620$ & partial---stability only\\
Pythia-70M & \texttt{direct\_logit\_only} & 30 & 512 & $0.864$ & $0.807$ & $9.00 \times 10^{-6}$ & $0.708$ & $0.694$ & $0.420$ & partial---stability only\\
Pythia-70M & \texttt{geometry\_only} & 30 & 512 & $0.843$ & $0.815$ & $0.028$ & $0.698$ & $0.733$ & $0.120$ & partial---stability only\\
Gemma-2-2B & \texttt{full\_no\_magnitude} & 25 & 256 & $0.174$ & $0.214$ & $0.130$ & $0.433$ & $0.450$ & $0.0018$ & partial---collateral only\\
Gemma-2-2B & \texttt{geometry\_only} & 25 & 256 & $0.196$ & $0.212$ & $0.850$ & $0.440$ & $0.444$ & $0.270$ & weak\\
Llama-3.1-8B & \texttt{full\_no\_magnitude} & 25 & 256 & $0.220$ & $0.212$ & $0.277$ & $0.209$ & $0.281$ & $1.32 \times 10^{-4}$ & partial---collateral only\\
Llama-3.1-8B & \texttt{geometry\_only} & 25 & 256 & $0.216$ & $0.209$ & $0.332$ & $0.221$ & $0.267$ & $0.0397$ & partial---collateral only\\
Llama-3.1-8B & \texttt{direct\_logit\_only} & 25 & 256 & $0.219$ & $0.205$ & $0.068$ & $0.236$ & $0.256$ & $0.332$ & weak\\
\hline
\end{tabular}%
}
\end{center}
\end{table}

\subsection{Held-out screening performance}

\begin{figure}[t]
\centering
\includegraphics[width=0.95\linewidth]{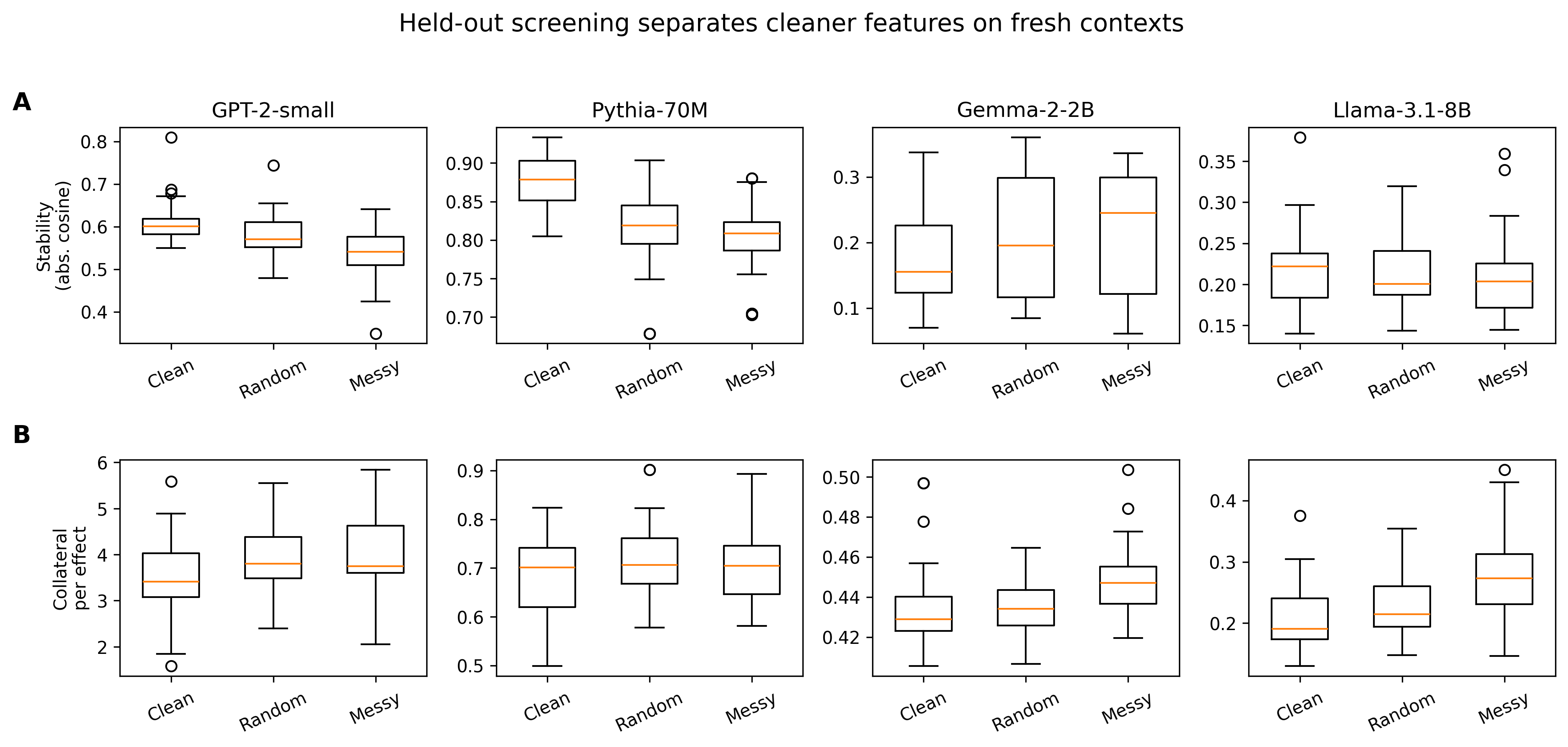}
\caption{Held-out screening distributions. Boxplots of predicted-clean, random-control, and predicted-messy feature groups on fresh contexts. Panel A shows absolute stability; Panel B shows collateral per effect.}
\label{fig:heldout-screening-distributions}
\end{figure}

The screening evaluation compared predicted-clean, predicted-messy, and random-control feature groups, selected out-of-sample and matched on predicted effect size, then steered on fresh held-out contexts. Group differences were assessed with two-sided Mann--Whitney tests (Table~\ref{tab:heldout-screening-performance}; distributions in Figure~\ref{fig:heldout-screening-distributions}). We report multiple-comparison-adjusted screening p-values and realized effect-size balance in Appendix Tables~\ref{tab:screening-multiple-corrections} and~\ref{tab:effect-size-balance}. These checks are not used to define new screening criteria, but they identify which clean--messy contrasts remain robust after correction and which may be affected by realized intervention magnitude.

In GPT-2-small (geometry-only selector, 32 features per group), predicted-clean features were more stable than predicted-messy features ($\Delta = +0.071$, $p = 4 \times 10^{-6}$) and lower in collateral per effect ($\Delta = -0.523$, $p = 0.020$); the two groups did not differ significantly in effect $L_2$ ($p = 0.12$). The contrast against the random control was also significant for stability ($\Delta = +0.032$, $p = 0.004$). Under the pre-specified raw-p-value rule this result is strong, with significant differences on both axes. The stability contrast survives all multiple-comparison corrections, while the collateral contrast remains significant under BH correction over the pre-specified main-selector contrasts but not under the stricter Holm correction; we therefore treat GPT-2 as the cleanest screening result, with stronger evidence for stability than for collateral. A GPT-2 intervention-strength sweep further shows that the held-out stability advantage is robust across $\alpha \in \{0.5,1.0,2.0\}$, while the collateral-per-effect advantage holds at $\alpha=0.5$ and $\alpha=1.0$ but attenuates at $\alpha=2.0$; effect $L_2$ remains balanced throughout (Appendix Table~\ref{tab:gpt2-alpha-sweep}).

In Pythia-70M, the stability contrast was significant under all three selectors (full no-magnitude $\Delta = +0.076$, $p = 1 \times 10^{-8}$; direct-logit-only $\Delta = +0.057$, $p = 9 \times 10^{-6}$; geometry-only $\Delta = +0.029$, $p = 0.028$), and the full no-magnitude stability difference was the largest single significant effect observed in the screening evaluation. The main full no-magnitude stability result survives multiple-comparison correction. The collateral-per-effect contrast was not significant for any selector (full no-magnitude $\Delta = -0.025$, $p = 0.62$). For the full no-magnitude and direct-logit selectors, the predicted-clean group additionally had a significantly larger effect $L_2$ than the predicted-messy group ($p < 0.01$), and the effect-size balance table confirms this imbalance for the main selector. We therefore interpret Pythia as strong evidence for stability screening, but not as clean evidence for low-collateral screening.

In Gemma-2-2B (25 features per group), the absolute-stability contrast was not significant and was negative under both selectors (full no-magnitude $\Delta = -0.040$, $p = 0.13$; geometry-only $\Delta = -0.016$, $p = 0.85$). The full no-magnitude selector produced a significant reduction in collateral per effect ($\Delta = -0.017$, $p = 0.0018$), whereas the geometry-only selector did not ($\Delta = -0.004$, $p = 0.27$). The collateral contrast survives multiple-comparison correction, but the effect-size balance check shows a small significant effect-magnitude difference between predicted-clean and predicted-messy groups. We therefore treat the Gemma result as partial and cautionary: it supports collateral-side screening, but not stability screening, and the collateral effect may be partly magnitude-sensitive.

In Llama-3.1-8B (25 features per group), held-out screening succeeded on the collateral axis but not the stability axis. Under the full no-magnitude selector, predicted-clean features had significantly lower collateral per effect than predicted-messy features ($0.209$ vs. $0.281$, $\Delta = -0.072$, $p = 1.3 \times 10^{-4}$), while the stability contrast was directionally positive but not significant ($0.220$ vs. $0.212$, $\Delta = +0.009$, $p = 0.277$). The two groups did not differ significantly in effect $L_2$ ($12.82$ vs. $12.85$, $p = 0.742$), indicating that the collateral reduction was not explained by realized intervention magnitude. This full no-magnitude collateral result survives multiple-comparison correction and is effect-size balanced, making it the clearest Llama screening finding. The geometry-only selector showed a directionally similar collateral-side pattern ($0.221$ vs. $0.267$, $\Delta = -0.046$, $p = 0.040$), but this auxiliary-selector effect does not survive correction. The direct-logit-only selector was directionally favorable for stability ($p = 0.068$) but did not reach significance on either pre-specified axis. Under the pre-specified rule, the full no-magnitude Llama result is partial, significant on collateral only.

\FloatBarrier

\section{Analysis and Discussion}

\subsection{Evidence by model/SAE setting}

The central question is whether pre-intervention SAE predictors can forecast feature-steering side effects. The results support this claim to different degrees across the four model/SAE settings, separating a broad claim about the \emph{method} from a narrower claim about any specific \emph{predictor}.

GPT-2-small provides the cleanest evidence because its predictive and screening results align. Decoder crowding had the strongest univariate relationship with both stability and collateral (Table~\ref{tab:dominant-predictor-by-model}); the no-magnitude predictors retained their advantage after residualizing out effect magnitude (Section~\ref{sec:residual}); and held-out screening improved both stability and collateral on fresh contexts without a significant effect-size difference (Table~\ref{tab:heldout-screening-performance}). Because prediction, robustness, and screening all point in the same direction, GPT-2 requires the fewest auxiliary assumptions to support the central claim.

Pythia-70M strengthens the case that the screening signal is not GPT-2-specific, especially for stability. Although Pythia differs from GPT-2 in both architecture and SAE setting, its predictive improvement over the activation-magnitude baseline was larger than GPT-2's ($+0.299$ versus $+0.249$), survived residualization, and produced a significant held-out stability contrast under all three selectors (Table~\ref{tab:heldout-screening-performance}). The collateral contrast was directionally favorable for two of three selectors but not statistically significant, and some selectors produced effect-size differences between clean and messy groups; we therefore treat Pythia collateral as unconfirmed. Still, the repeated predictive and stability-screening signal suggests that the predictor--modularity relationship is not idiosyncratic to GPT-2.

Llama-3.1-8B provides the strongest larger-model evidence for prediction. The no-magnitude predictor set produced the largest mean improvements over the frequency and activation-magnitude baselines among the evaluated settings ($+0.451$ and $+0.386$), and the residualized-stability signal remained nontrivial after controlling for magnitude-related variables (Section~\ref{sec:residual}; Appendix~\ref{app:llama}). In held-out screening, however, the benefit appeared on collateral rather than stability: full no-magnitude predicted-clean features had significantly lower collateral per effect than predicted-messy features, with no significant effect-magnitude difference, but the stability contrast was not significant (Table~\ref{tab:heldout-screening-performance}). Llama therefore strengthens the method-level claim in a modern 8B model while showing that the successful screening axis can change across dictionary settings.

Gemma-2-2B shows the weakest transfer. Its predictive gains were genuine but smaller ($+0.193$ over frequency, compared with $+0.390$, $+0.421$, and $+0.451$ in GPT-2, Pythia, and Llama), and they did not survive residualization as cleanly, falling near zero on absolute stability. In screening, Gemma showed no stability benefit and only an isolated reduction in collateral per effect under one selector (Table~\ref{tab:heldout-screening-performance}). The Gemma evidence is therefore consistent with a weaker, more magnitude-entangled version of the predictor--modularity relationship.

After multiple-comparison correction and effect-size-balance checks, the held-out screening evidence remains real but axis-specific. GPT-2 gives the cleanest overall screening result, with stability surviving all corrections and collateral remaining supportive but less robust under strict Holm correction. Pythia gives strong corrected evidence for stability screening, although the main selector is not effect-size balanced. Llama gives the clearest corrected and effect-size-balanced collateral result. Gemma gives corrected collateral evidence, but the effect is small and accompanied by slight effect-size imbalance. Thus, the screening results support the practical value of pre-intervention ranking, but only in a setting- and axis-dependent sense.

\subsection{Method-level versus mechanism-level transfer}

Across the four model/SAE settings, the most informative predictor changes (Table~\ref{tab:dominant-predictor-by-model}; Figure~\ref{fig:representative-univariate-relationships.}). GPT-2 is dominated by decoder geometry, especially decoder crowding. Pythia relies more on broader activation-independent signals and direct-logit footprint, with geometry alone predicting collateral poorly. Llama shows strong geometry and direct-logit signal, especially for collateral and effect-related targets. Gemma shows weaker geometry-only transfer, with encoder-scale geometry leading.

To contextualize this setting dependence, we computed dictionary-level diagnostics over the same 300 sampled features used in the predictive evaluation (Appendix Table~\ref{tab:dictionary-diagnostics}). The four settings occupy different geometric and sparsity regimes. GPT-2-small and Pythia-70M have the highest mean decoder crowding, consistent with decoder geometry being informative in the ReLU-SAE settings. Gemma-2-2B has much lower decoder crowding but a much larger coactivation count, suggesting a different downstream activation regime under the JumpReLU dictionary. Llama-3.1-8B has meaningful decoder-norm variation and the largest direct-logit footprint dispersion, consistent with its strong geometry and direct-logit signals. Because this analysis compares only four settings, we treat these diagnostics as explanatory context rather than a causal test. They nevertheless suggest that setting-dependent predictor signatures are not arbitrary: they track measurable differences in SAE dictionary geometry and sparsity.

The controlled Llama width comparison further supports this distinction: pre-intervention predictability persists under a 32K-to-128K dictionary-width change, but the held-out screening payoff becomes less stable and shifts toward a direct-logit stability-only effect. Thus, the general predictive method is more robust than the particular screening axis or predictor signature that transfers in a given dictionary.

This pattern motivates a distinction between \emph{method-level transfer} and \emph{mechanism-level transfer}. Method-level transfer is the claim that fitting cheap, intervention-free predictors and using them to screen features continues to outperform naive baselines in new model/SAE settings. This claim is supported across the four settings, but not identically: GPT-2 shows both-axis screening, Pythia shows stability screening, Llama shows collateral screening, and Gemma shows the weakest and most magnitude-entangled transfer. Mechanism-level transfer is the stronger claim that a specific predictor signature, such as decoder crowding, governs side effects across settings; this did not hold. The dominant signal changes across models and dictionaries, and the transferred modularity axis also changes. Thus, the results show that steering side effects carry a cheap, pre-intervention signature, not that any single feature property or universal notion of cleanliness transfers across all settings.

\section{Conclusion}
\label{sec:conclusion}

We asked whether SAE steering side effects can be forecast before intervention from cheap feature statistics. Across four model/SAE settings, intervention-free predictors recovered steering modularity and improved over frequency-only and activation-magnitude baselines, with the strongest and most robust evidence in GPT-2-small, Pythia-70M, and Llama-3.1-8B, and weaker, more magnitude-entangled evidence in Gemma-2-2B. Held-out screening further showed that predicted-clean features can steer more cleanly on fresh contexts, but the successful axis differs across settings: both stability and collateral in GPT-2, stability in Pythia, collateral in Llama, and partial collateral evidence in Gemma.

The central conclusion is therefore bounded: SAE steering side effects are predictable in advance, but the predictive signature and practical screening payoff are model- and dictionary-setting dependent rather than universal. Several limitations qualify this conclusion, including small held-out screening groups, a single main intervention strength for the primary cross-model experiments, with only a limited GPT-2 sensitivity sweep, collateral measurement through a downstream SAE proxy, and incomplete disentanglement of model and SAE-family effects. Future work should vary SAE width and family within fixed base models, enlarge screening pools, and connect pre-intervention screening to end-to-end steering pipelines where cleaner feature selection improves targeted control.

\section{Acknowledgements}

All research questions, experimental design choices, implementation decisions, numerical results, analyses, and final claims were developed, reviewed, and verified by the authors.

%% file: appendix_table_a1.tex
\section{Experimental details}

This appendix reports the full experimental configuration for the original three non-Llama settings. Llama-3.1-8B details are reported separately in Appendix~\ref{app:llama}. Table A1 lists the model checkpoints, SAE releases, and primary/downstream hook sites. Table A2 reports the predictive-evaluation configuration, including corpus, context, feature-sampling, and intervention settings. Table A3 reports the held-out screening configuration, including selector sets, group sizes, and fresh-context evaluation scale.
 
\begin{table}[!htbp]
\caption{Model and SAE configuration. Exact model checkpoints, SAE releases, and primary/downstream hook sites used for each model replication.}
\label{tab:appendix-model-sae-config}
\begin{center}
\scriptsize
\setlength{\tabcolsep}{3pt}
\resizebox{\linewidth}{!}{%
\begin{tabular}{@{}lllll@{}}
\hline
\textbf{Model} & \textbf{Model checkpoint} & \textbf{SAE release} & \textbf{Primary SAE / hook} & \textbf{Downstream SAE / hook}\\
\hline
GPT-2-small & \texttt{gpt2-small} & \texttt{gpt2-small-res-jb} & \texttt{blocks.8.hook\_resid\_pre} & \texttt{blocks.10.hook\_resid\_pre}\\
Pythia-70M-deduped & \texttt{EleutherAI/pythia-70m-deduped} & \texttt{pythia-70m-deduped-res-sm} & \texttt{blocks.4.hook\_resid\_post} & \texttt{blocks.5.hook\_resid\_post}\\
Gemma-2-2B & \texttt{gemma-2-2b} & \texttt{gemma-scope-2b-pt-res-canonical} & \texttt{layer\_12/width\_16k/canonical} & \texttt{layer\_16/width\_16k/canonical}\\
\hline
\end{tabular}%
}
\end{center}
\end{table}

\begin{table}[!htbp]
\caption{Predictive-evaluation configuration. Corpus, context, feature-sampling, and primary intervention settings used to construct steering labels and predictive-evaluation targets.}
\label{tab:appendix-predictive-evaluation-config}
\begin{center}
\scriptsize
\setlength{\tabcolsep}{3pt}
\resizebox{\linewidth}{!}{%
\begin{tabular}{@{}llllllllll@{}}
\hline
\textbf{Model} & \textbf{Dataset} & \textbf{Split} & \textbf{Distinct texts} & \textbf{Context length} & \textbf{Features} & \textbf{Contexts} & \textbf{Contexts/type} & \textbf{Primary intervention} & \textbf{Clamp/add value}\\
\hline
GPT-2-small & Wikitext-103 & \texttt{train[:2\%]} & 8,000 & 48 & 300 & 2,048 & 16 & \texttt{fixed\_global\_add} & 1.0\\
Pythia-70M-deduped & Wikitext-103 & \texttt{train[:2\%]} & 8,000 & 48 & 300 & 2,048 & 16 & \texttt{fixed\_global\_add} & 1.0\\
Gemma-2-2B & Wikitext-103 & \texttt{train[:2\%]} & 8,000 & 48 & 300 & 2,048 & 16 & \texttt{fixed\_global\_add} & 1.0\\
\hline
\end{tabular}%
}
\end{center}
\end{table}

\begin{table}[!htbp]
\caption{Held-out screening configuration. Feature-selection rules, group sizes, and fresh-context evaluation settings used in the held-out screening experiments.}
\label{tab:appendix-heldout-screening-config}
\begin{center}
\scriptsize
\setlength{\tabcolsep}{3pt}
\resizebox{\linewidth}{!}{%
\begin{tabular}{@{}lllll@{}}
\hline
\textbf{Model} & \textbf{Screening selectors} & \textbf{$n$/group} & \textbf{Held-out contexts} & \textbf{Main reported selector}\\
\hline
GPT-2-small & \texttt{geometry\_only} & 32 & 512 & \texttt{geometry\_only}\\
Pythia-70M-deduped & \texttt{full\_no\_magnitude; direct\_logit\_only; geometry\_only} & 30 & 512 & \texttt{full\_no\_magnitude}\\
Gemma-2-2B & \texttt{full\_no\_magnitude; geometry\_only} & 25 & 256 & \texttt{full\_no\_magnitude}\\
\hline
\end{tabular}%
}
\end{center}
\end{table}

%% file: appendix_predictive_evaluation.tex
\section{Predictive Evaluation Details}

This appendix provides the full predictive-evaluation results underlying the main-text summary. Table B1 reports cross-validated Spearman performance for all predictor sets and steering-label targets. Table B2 reports the full univariate predictor-label correlations. Table B3 reports the residualized-target robustness analysis after controlling for effect magnitude, intervention value, and natural feature activation.

Llama results are reported separately in Appendix~\ref{app:llama}.

\begin{table}[!htbp]
\caption{Full predictive baseline comparison. Cross-validated Spearman performance for all predictor sets and primary steering-label targets across GPT-2-small, Pythia-70M, and Gemma-2-2B. Corresponding file: \texttt{appendix\_\allowbreak{}table\_\allowbreak{}A2\_\allowbreak{}full\_\allowbreak{}baseline\_\allowbreak{}comparison.csv}.}
\label{tab:appendix-baseline-comparison}
\begin{center}
\tiny
\setlength{\tabcolsep}{2pt}
\resizebox{\linewidth}{!}{%
\begin{tabular}{@{}llllllllllllll@{}}
\hline
\textbf{Model} & \textbf{Target} & \textbf{Freq.} & \textbf{Act. mag.} & \textbf{Crowding} & \textbf{Geom.} & \textbf{Coact.} & \textbf{Direct logit} & \textbf{Compact no-mag} & \textbf{Full no-mag} & \textbf{Full all} & \textbf{$\Delta$ freq.} & \textbf{$\Delta$ act.} & \textbf{$\Delta$ all-act.}\\
\hline
GPT-\allowbreak{}2-\allowbreak{}small & abs stability & $0.118$ & $0.248$ & $0.361$ & $0.516$ & $0.0311$ & $0.324$ & $0.45$ & $0.552$ & $0.563$ & $0.434$ & $0.305$ & $0.315$\\
GPT-\allowbreak{}2-\allowbreak{}small & signed stability & $0.119$ & $0.25$ & $0.362$ & $0.517$ & $0.0334$ & $0.326$ & $0.451$ & $0.554$ & $0.565$ & $0.435$ & $0.304$ & $0.315$\\
GPT-\allowbreak{}2-\allowbreak{}small & pairwise abs cosine & $0.0879$ & $0.228$ & $0.35$ & $0.509$ & $0.0187$ & $0.316$ & $0.439$ & $0.559$ & $0.565$ & $0.471$ & $0.33$ & $0.336$\\
GPT-\allowbreak{}2-\allowbreak{}small & pairwise signed cosine & $0.0996$ & $0.239$ & $0.356$ & $0.515$ & $0.0252$ & $0.324$ & $0.445$ & $0.563$ & $0.569$ & $0.464$ & $0.324$ & $0.33$\\
GPT-\allowbreak{}2-\allowbreak{}small & effect CV & $-0.0397$ & $0.0552$ & $0.198$ & $0.208$ & $0.104$ & $0.156$ & $0.133$ & $0.127$ & $0.117$ & $0.167$ & $0.0721$ & $0.0618$\\
GPT-\allowbreak{}2-\allowbreak{}small & downstream $L_2$ & $0.216$ & $0.404$ & $0.419$ & $0.553$ & $0.0766$ & $0.0847$ & $0.566$ & $0.622$ & $0.634$ & $0.407$ & $0.218$ & $0.23$\\
GPT-\allowbreak{}2-\allowbreak{}small & effective moved downstream & $-0.00576$ & $-0.00733$ & $0.0353$ & $-0.102$ & $0.209$ & $0.103$ & $0.188$ & $0.177$ & $0.239$ & $0.182$ & $0.184$ & $0.246$\\
GPT-\allowbreak{}2-\allowbreak{}small & downstream count $>0.05$ & $-0.000104$ & $0.301$ & $0.433$ & $0.512$ & $0.202$ & $0.0561$ & $0.507$ & $0.556$ & $0.596$ & $0.557$ & $0.255$ & $0.295$\\
Pythia-\allowbreak{}70M & abs stability & $0.0946$ & $0.292$ & $0.352$ & $0.389$ & $0.156$ & $0.428$ & $0.55$ & $0.59$ & $0.585$ & $0.495$ & $0.298$ & $0.294$\\
Pythia-\allowbreak{}70M & signed stability & $0.0946$ & $0.292$ & $0.352$ & $0.389$ & $0.156$ & $0.428$ & $0.55$ & $0.59$ & $0.585$ & $0.495$ & $0.298$ & $0.294$\\
Pythia-\allowbreak{}70M & pairwise abs cosine & $0.0967$ & $0.293$ & $0.351$ & $0.392$ & $0.169$ & $0.42$ & $0.546$ & $0.589$ & $0.583$ & $0.492$ & $0.296$ & $0.29$\\
Pythia-\allowbreak{}70M & pairwise signed cosine & $0.0969$ & $0.293$ & $0.351$ & $0.392$ & $0.17$ & $0.42$ & $0.546$ & $0.589$ & $0.583$ & $0.492$ & $0.296$ & $0.29$\\
Pythia-\allowbreak{}70M & effect CV & $0.0406$ & $0.154$ & $-0.0831$ & $0.0376$ & $0.109$ & $0.312$ & $0.281$ & $0.296$ & $0.284$ & $0.256$ & $0.142$ & $0.13$\\
Pythia-\allowbreak{}70M & downstream $L_2$ & $0.0659$ & $4.04e-05$ & $0.116$ & $0.0521$ & $0.0942$ & $0.266$ & $0.286$ & $0.293$ & $0.292$ & $0.227$ & $0.293$ & $0.292$\\
Pythia-\allowbreak{}70M & effective moved downstream & $-0.023$ & $0.0509$ & $-0.019$ & $-0.0236$ & $0.278$ & $0.139$ & $0.232$ & $0.222$ & $0.313$ & $0.245$ & $0.171$ & $0.262$\\
Pythia-\allowbreak{}70M & downstream count $>0.05$ & $-0.0495$ & $0.0192$ & $-0.0537$ & $0.0867$ & $0.19$ & $0.588$ & $0.591$ & $0.613$ & $0.588$ & $0.663$ & $0.594$ & $0.569$\\
Gemma-\allowbreak{}2-\allowbreak{}2B & abs stability & $0.031$ & $-0.181$ & $-0.0557$ & $-0.035$ & $-0.0724$ & $0.118$ & $0.0521$ & $0.0276$ & $0.00441$ & $-0.00344$ & $0.208$ & $0.185$\\
Gemma-\allowbreak{}2-\allowbreak{}2B & signed stability & $0.0892$ & $0.119$ & $0.306$ & $0.293$ & $0.127$ & $0.375$ & $0.387$ & $0.396$ & $0.382$ & $0.307$ & $0.277$ & $0.264$\\
Gemma-\allowbreak{}2-\allowbreak{}2B & pairwise abs cosine & $-0.0345$ & $-0.144$ & $-0.0234$ & $-0.0469$ & $0.104$ & $-0.112$ & $-0.0107$ & $0.0191$ & $0.00874$ & $0.0535$ & $0.163$ & $0.152$\\
Gemma-\allowbreak{}2-\allowbreak{}2B & pairwise signed cosine & $-0.0936$ & $0.103$ & $0.262$ & $0.281$ & $-0.0443$ & $0.348$ & $0.328$ & $0.327$ & $0.319$ & $0.42$ & $0.224$ & $0.216$\\
Gemma-\allowbreak{}2-\allowbreak{}2B & effect CV & $0.0993$ & $0.0216$ & $-0.0209$ & $-0.0538$ & $0.123$ & $-0.0434$ & $0.0954$ & $0.0856$ & $0.047$ & $-0.0137$ & $0.064$ & $0.0253$\\
Gemma-\allowbreak{}2-\allowbreak{}2B & downstream $L_2$ & $-0.0922$ & $-0.0356$ & $0.121$ & $0.188$ & $0.0169$ & $0.0337$ & $0.221$ & $0.219$ & $0.264$ & $0.311$ & $0.255$ & $0.3$\\
Gemma-\allowbreak{}2-\allowbreak{}2B & effective moved downstream & $-0.0246$ & $0.029$ & $0.123$ & $0.197$ & $0.2$ & $0.0545$ & $0.212$ & $0.203$ & $0.225$ & $0.228$ & $0.174$ & $0.196$\\
Gemma-\allowbreak{}2-\allowbreak{}2B & downstream count $>0.05$ & $0.138$ & $0.293$ & $0.172$ & $0.329$ & $0.107$ & $-0.0388$ & $0.355$ & $0.381$ & $0.391$ & $0.243$ & $0.0878$ & $0.0983$\\
\hline
\end{tabular}%
}
\end{center}
\end{table}

\begin{table}[!htbp]
\caption{Full univariate predictor correlations. Spearman and Pearson correlations between individual pre-intervention predictors and measured steering labels for all three model/SAE settings. Corresponding file: \texttt{appendix\_\allowbreak{}table\_\allowbreak{}A3\_\allowbreak{}full\_\allowbreak{}univariate\_\allowbreak{}correlations.csv}.}
\label{tab:appendix-univariate-correlations}
\begin{center}
\scriptsize
\setlength{\tabcolsep}{3pt}
\resizebox{\linewidth}{!}{%
\begin{tabular}{@{}lllllllll@{}}
\hline
\textbf{Model} & \textbf{Rows} & \textbf{Predictors} & \textbf{Targets} & \textbf{Primary rows} & \textbf{Strongest predictor} & \textbf{Strongest target} & \textbf{Spearman $r$} & \textbf{Pearson $r$}\\
\hline
GPT-\allowbreak{}2-\allowbreak{}small & 5040 & 28 & 180 & 224 & phase1\_\allowbreak{}final\_\allowbreak{}act\_\allowbreak{}max & effect\_\allowbreak{}cv\_\allowbreak{}\_\allowbreak{}fixed\_\allowbreak{}global\_\allowbreak{}\_\allowbreak{}mixed & $0.898$ & $0.887$\\
Gemma-\allowbreak{}2-\allowbreak{}2B & 5040 & 28 & 180 & 224 & phase1\_\allowbreak{}final\_\allowbreak{}act\_\allowbreak{}mean & downstream\_\allowbreak{}feat\_\allowbreak{}count\_\allowbreak{}abs\_\allowbreak{}delta\_\allowbreak{}gt\_\allowbreak{}0.1\_\allowbreak{}\_\allowbreak{}fixed\_\allowbreak{}global\_\allowbreak{}\_\allowbreak{}top & $0.925$ & $0.318$\\
Pythia-\allowbreak{}70M & 5040 & 28 & 180 & 224 & phase1\_\allowbreak{}final\_\allowbreak{}act\_\allowbreak{}max & stability\_\allowbreak{}to\_\allowbreak{}mean\_\allowbreak{}signed\_\allowbreak{}cos\_\allowbreak{}\_\allowbreak{}feature\_\allowbreak{}q95\_\allowbreak{}\_\allowbreak{}top & $-0.738$ & $-0.608$\\
\hline
\end{tabular}%
}
\end{center}
\end{table}

\begin{table}[!htbp]
\caption{Full residualized-target robustness results. Predictive performance after residualizing stability labels against effect magnitude, intervention value, and natural feature activation. Corresponding file: \texttt{appendix\_\allowbreak{}table\_\allowbreak{}A4\_\allowbreak{}full\_\allowbreak{}residualized\_\allowbreak{}results.csv}.}
\label{tab:appendix-residualized-results}
\begin{center}
\begingroup
\fontsize{6}{6.2}\selectfont
\renewcommand{\arraystretch}{0.78}
\setlength{\tabcolsep}{2pt}
\resizebox{\linewidth}{!}{%
\begin{tabular}{@{}llllllll@{}}
\hline
\textbf{Model} & \textbf{Original target} & \textbf{Predictor set} & \textbf{$n$} & \textbf{$p$} & \textbf{CV $R^2$} & \textbf{CV MAE} & \textbf{CV Spearman}\\
\hline
GPT-\allowbreak{}2-\allowbreak{}small & abs stability & geometry only & 300 & 6 & $0.184$ & $0.0494$ & $0.367$\\
GPT-\allowbreak{}2-\allowbreak{}small & abs stability & coactivation only & 300 & 2 & $-0.0117$ & $0.0513$ & $-0.00868$\\
GPT-\allowbreak{}2-\allowbreak{}small & abs stability & compact no-mag & 300 & 16 & $0.377$ & $0.0433$ & $0.511$\\
GPT-\allowbreak{}2-\allowbreak{}small & abs stability & full no-mag & 300 & 22 & $0.41$ & $0.041$ & $0.555$\\
GPT-\allowbreak{}2-\allowbreak{}small & abs stability & full all & 300 & 29 & $-91.8$ & $0.0812$ & $0.585$\\
GPT-\allowbreak{}2-\allowbreak{}small & signed stability & geometry only & 300 & 6 & $0.184$ & $0.0494$ & $0.369$\\
GPT-\allowbreak{}2-\allowbreak{}small & signed stability & coactivation only & 300 & 2 & $-0.0114$ & $0.0513$ & $-0.00607$\\
GPT-\allowbreak{}2-\allowbreak{}small & signed stability & compact no-mag & 300 & 16 & $0.378$ & $0.0433$ & $0.511$\\
GPT-\allowbreak{}2-\allowbreak{}small & signed stability & full no-mag & 300 & 22 & $0.411$ & $0.0409$ & $0.556$\\
GPT-\allowbreak{}2-\allowbreak{}small & signed stability & full all & 300 & 29 & $-92.7$ & $0.0813$ & $0.586$\\
GPT-\allowbreak{}2-\allowbreak{}small & pairwise abs cosine & geometry only & 300 & 6 & $0.189$ & $0.0544$ & $0.399$\\
GPT-\allowbreak{}2-\allowbreak{}small & pairwise abs cosine & coactivation only & 300 & 2 & $-0.00926$ & $0.0572$ & $-0.0177$\\
GPT-\allowbreak{}2-\allowbreak{}small & pairwise abs cosine & compact no-mag & 300 & 16 & $0.324$ & $0.0505$ & $0.457$\\
GPT-\allowbreak{}2-\allowbreak{}small & pairwise abs cosine & full no-mag & 300 & 22 & $0.377$ & $0.0472$ & $0.528$\\
GPT-\allowbreak{}2-\allowbreak{}small & pairwise abs cosine & full all & 300 & 29 & $-100$ & $0.0946$ & $0.554$\\
GPT-\allowbreak{}2-\allowbreak{}small & pairwise signed cosine & geometry only & 300 & 6 & $0.192$ & $0.0568$ & $0.394$\\
GPT-\allowbreak{}2-\allowbreak{}small & pairwise signed cosine & coactivation only & 300 & 2 & $-0.00996$ & $0.0597$ & $-0.00856$\\
GPT-\allowbreak{}2-\allowbreak{}small & pairwise signed cosine & compact no-mag & 300 & 16 & $0.341$ & $0.0521$ & $0.473$\\
GPT-\allowbreak{}2-\allowbreak{}small & pairwise signed cosine & full no-mag & 300 & 22 & $0.391$ & $0.0487$ & $0.538$\\
GPT-\allowbreak{}2-\allowbreak{}small & pairwise signed cosine & full all & 300 & 29 & $-104$ & $0.0995$ & $0.566$\\
Pythia-\allowbreak{}70M & abs stability & geometry only & 300 & 6 & $0.317$ & $0.0263$ & $0.499$\\
Pythia-\allowbreak{}70M & abs stability & coactivation only & 300 & 2 & $0.0534$ & $0.0305$ & $0.274$\\
Pythia-\allowbreak{}70M & abs stability & compact no-mag & 300 & 16 & $0.229$ & $0.0278$ & $0.416$\\
Pythia-\allowbreak{}70M & abs stability & full no-mag & 300 & 22 & $0.31$ & $0.0265$ & $0.49$\\
Pythia-\allowbreak{}70M & abs stability & full all & 300 & 29 & $0.131$ & $0.0272$ & $0.489$\\
Pythia-\allowbreak{}70M & signed stability & geometry only & 300 & 6 & $0.317$ & $0.0263$ & $0.499$\\
Pythia-\allowbreak{}70M & signed stability & coactivation only & 300 & 2 & $0.0534$ & $0.0305$ & $0.274$\\
Pythia-\allowbreak{}70M & signed stability & compact no-mag & 300 & 16 & $0.229$ & $0.0278$ & $0.416$\\
Pythia-\allowbreak{}70M & signed stability & full no-mag & 300 & 22 & $0.31$ & $0.0265$ & $0.49$\\
Pythia-\allowbreak{}70M & signed stability & full all & 300 & 29 & $0.131$ & $0.0272$ & $0.489$\\
Pythia-\allowbreak{}70M & pairwise abs cosine & geometry only & 300 & 6 & $0.326$ & $0.0443$ & $0.496$\\
Pythia-\allowbreak{}70M & pairwise abs cosine & coactivation only & 300 & 2 & $0.0591$ & $0.0516$ & $0.274$\\
Pythia-\allowbreak{}70M & pairwise abs cosine & compact no-mag & 300 & 16 & $0.237$ & $0.0471$ & $0.416$\\
Pythia-\allowbreak{}70M & pairwise abs cosine & full no-mag & 300 & 22 & $0.321$ & $0.0445$ & $0.49$\\
Pythia-\allowbreak{}70M & pairwise abs cosine & full all & 300 & 29 & $0.135$ & $0.0458$ & $0.49$\\
Pythia-\allowbreak{}70M & pairwise signed cosine & geometry only & 300 & 6 & $0.326$ & $0.0443$ & $0.496$\\
Pythia-\allowbreak{}70M & pairwise signed cosine & coactivation only & 300 & 2 & $0.059$ & $0.0516$ & $0.275$\\
Pythia-\allowbreak{}70M & pairwise signed cosine & compact no-mag & 300 & 16 & $0.237$ & $0.0471$ & $0.416$\\
Pythia-\allowbreak{}70M & pairwise signed cosine & full no-mag & 300 & 22 & $0.321$ & $0.0445$ & $0.49$\\
Pythia-\allowbreak{}70M & pairwise signed cosine & full all & 300 & 29 & $0.136$ & $0.0458$ & $0.491$\\
Gemma-\allowbreak{}2-\allowbreak{}2B & abs stability & geometry only & 300 & 6 & $-0.0406$ & $0.0675$ & $-0.0847$\\
Gemma-\allowbreak{}2-\allowbreak{}2B & abs stability & coactivation only & 300 & 2 & $-0.0368$ & $0.067$ & $-0.18$\\
Gemma-\allowbreak{}2-\allowbreak{}2B & abs stability & compact no-mag & 300 & 16 & $-0.133$ & $0.0683$ & $0.0335$\\
Gemma-\allowbreak{}2-\allowbreak{}2B & abs stability & full no-mag & 300 & 22 & $-0.154$ & $0.0689$ & $0.00835$\\
Gemma-\allowbreak{}2-\allowbreak{}2B & abs stability & full all & 300 & 29 & $-9.99$ & $0.0826$ & $0.0136$\\
Gemma-\allowbreak{}2-\allowbreak{}2B & signed stability & geometry only & 300 & 6 & $-0.0432$ & $0.0244$ & $0.0651$\\
Gemma-\allowbreak{}2-\allowbreak{}2B & signed stability & coactivation only & 300 & 2 & $-0.00701$ & $0.0243$ & $0.018$\\
Gemma-\allowbreak{}2-\allowbreak{}2B & signed stability & compact no-mag & 300 & 16 & $-0.159$ & $0.0244$ & $0.222$\\
Gemma-\allowbreak{}2-\allowbreak{}2B & signed stability & full no-mag & 300 & 22 & $-0.171$ & $0.0248$ & $0.237$\\
Gemma-\allowbreak{}2-\allowbreak{}2B & signed stability & full all & 300 & 29 & $-18.5$ & $0.0317$ & $0.258$\\
Gemma-\allowbreak{}2-\allowbreak{}2B & pairwise abs cosine & geometry only & 300 & 6 & $0.018$ & $0.015$ & $0.173$\\
Gemma-\allowbreak{}2-\allowbreak{}2B & pairwise abs cosine & coactivation only & 300 & 2 & $-0.00151$ & $0.0153$ & $0.0486$\\
Gemma-\allowbreak{}2-\allowbreak{}2B & pairwise abs cosine & compact no-mag & 300 & 16 & $0.0174$ & $0.015$ & $0.222$\\
Gemma-\allowbreak{}2-\allowbreak{}2B & pairwise abs cosine & full no-mag & 300 & 22 & $0.0202$ & $0.015$ & $0.219$\\
Gemma-\allowbreak{}2-\allowbreak{}2B & pairwise abs cosine & full all & 300 & 29 & $-4.92$ & $0.0175$ & $0.211$\\
Gemma-\allowbreak{}2-\allowbreak{}2B & pairwise signed cosine & geometry only & 300 & 6 & $-0.0564$ & $0.00806$ & $0.0338$\\
Gemma-\allowbreak{}2-\allowbreak{}2B & pairwise signed cosine & coactivation only & 300 & 2 & $-0.0184$ & $0.00802$ & $-0.0964$\\
Gemma-\allowbreak{}2-\allowbreak{}2B & pairwise signed cosine & compact no-mag & 300 & 16 & $0.0009$ & $0.0079$ & $0.188$\\
Gemma-\allowbreak{}2-\allowbreak{}2B & pairwise signed cosine & full no-mag & 300 & 22 & $-0.0138$ & $0.00804$ & $0.195$\\
Gemma-\allowbreak{}2-\allowbreak{}2B & pairwise signed cosine & full all & 300 & 29 & $-8.75$ & $0.00982$ & $0.196$\\
\hline
\end{tabular}%
}
\endgroup
\end{center}
\end{table}

%% file: appendix_predictor_label_diagnostics.tex
\section{Predictor and Label Diagnostics}

\begin{figure}[!htbp]
\centering
\includegraphics[width=0.72\linewidth,height=0.38\textheight,keepaspectratio]{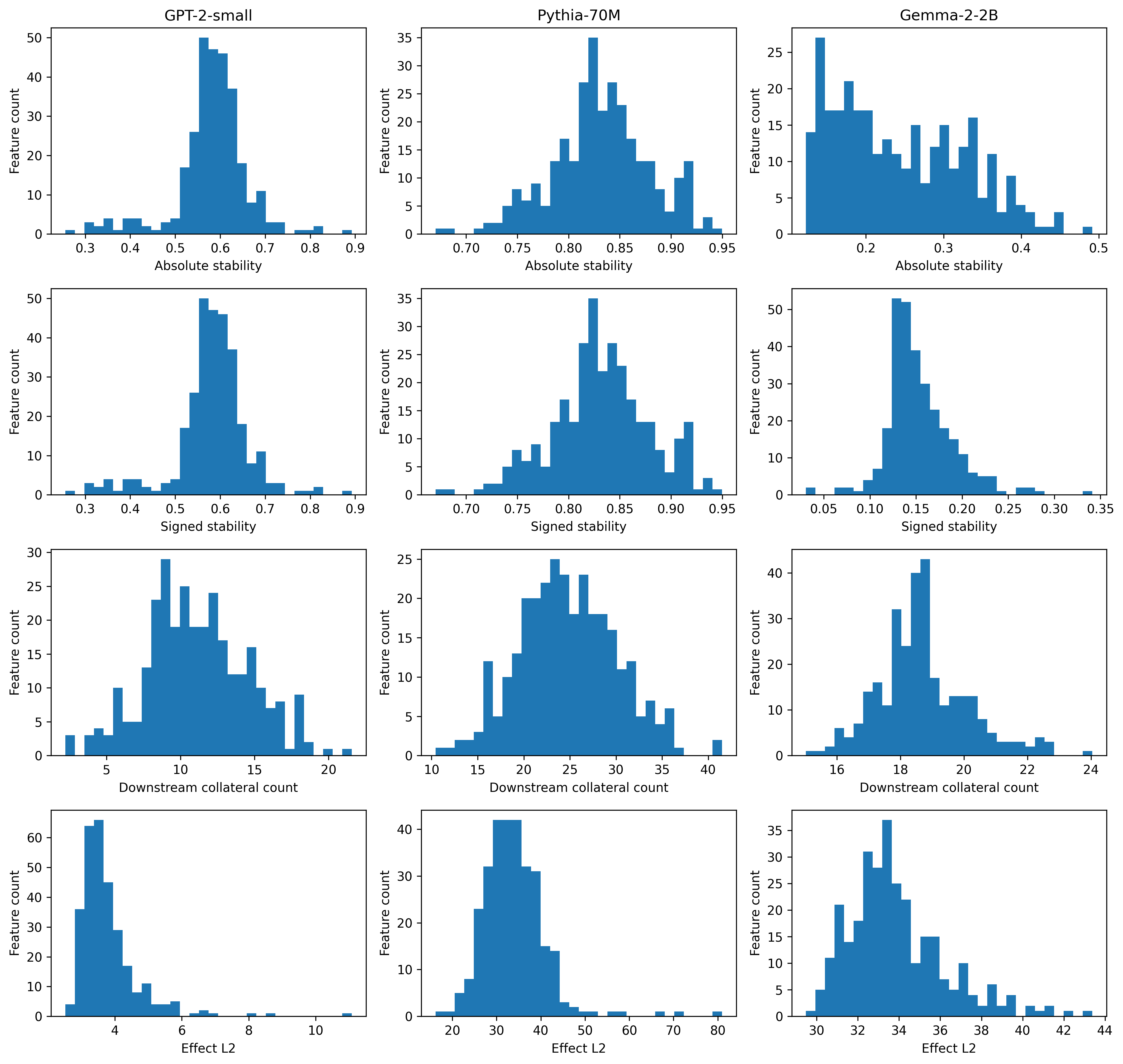}
\caption[Label distributions across models]{Label distributions across models. Distributions of predictive-evaluation steering labels for GPT-2-small, Pythia-70M, and Gemma-2-2B, including stability, downstream collateral count, and effect magnitude. Corresponding files: \protect\path{figure_C1_label_distributions.png} and \protect\path{figure_C1_label_distributions.pdf}.}
\label{fig:appendix-label-distributions}

\vspace{10pt}

\includegraphics[width=0.80\linewidth,height=0.24\textheight,keepaspectratio]{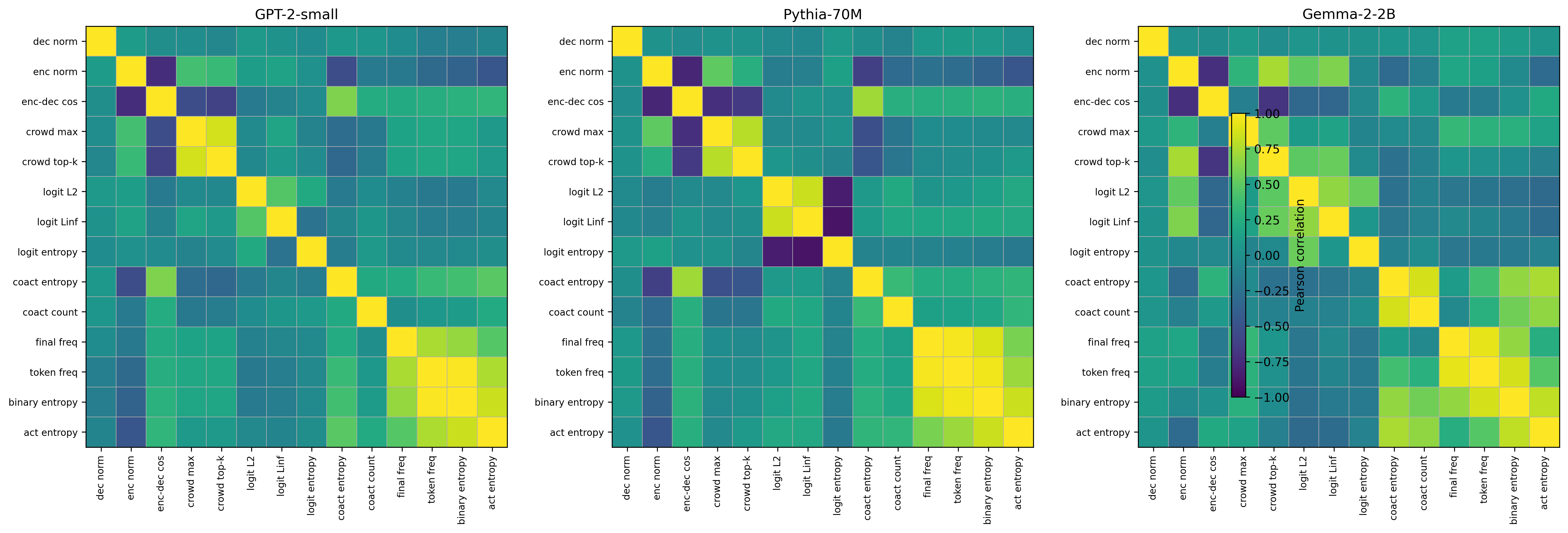}
\caption[Predictor correlation structure]{Predictor correlation structure. Pearson correlation heatmaps among intervention-free predictors for each model/SAE setting, showing redundancy among geometry, activation, co-activation, and direct-logit predictor families. Corresponding files: \protect\path{figure_C2_predictor_correlation_heatmaps.png} and \protect\path{figure_C2_predictor_correlation_heatmaps.pdf}.}
\label{fig:appendix-predictor-correlation-heatmaps}
\end{figure}

This appendix provides diagnostic plots for the feature-level labels and intervention-free predictors used in the predictive evaluation. Table~\ref{tab:dictionary-diagnostics} summarizes dictionary-level diagnostics across model/SAE settings. Figure~\ref{fig:appendix-label-distributions} shows the distributions of the main steering-label targets across models, verifying that the prediction targets have non-degenerate variation. Figure~\ref{fig:appendix-predictor-correlation-heatmaps} shows correlations among predictor families, illustrating redundancy among geometry, activation, co-activation, and direct-logit statistics. Llama-specific diagnostics and width-control results are reported separately in Appendix~\ref{app:llama-width-control}.

\begin{table}[!htbp]
\caption{Dictionary-level diagnostics across model/SAE settings. Summary statistics are computed over the same 300 sampled SAE features used in the predictive evaluation. The diagnostics show substantial variation in decoder crowding, norm dispersion, coactivation density, and direct-logit footprint dispersion across SAE families. The final columns report the dominant side-effect-relevant predictor family and the held-out screening axis observed in the main analysis.}
\label{tab:dictionary-diagnostics}
\begin{center}
\small
\setlength{\tabcolsep}{3pt}
\resizebox{\textwidth}{!}{%
\begin{tabular}{@{}llrrrrrlll@{}}
\toprule
Model & SAE family & Crowd. & Dec. norm CV & Enc. norm CV & Freq. & Coact. count & Logit $L_2$ CV & Dominant family & Screening axis \\
\midrule
GPT-2-small & ReLU & 0.390 & 0.000 & 0.320 & 0.009 & 39.013 & 0.192 & geometry / crowding & stability + collateral \\
Pythia-70M & ReLU & 0.369 & 0.000 & 0.362 & 0.013 & 54.379 & 0.193 & direct-logit / crowding & stability \\
Gemma-2-2B & JumpReLU & 0.184 & 0.000 & 0.187 & 0.010 & 458.728 & 0.193 & encoder geometry & collateral weakly \\
Llama-3.1-8B & TopK & 0.258 & 0.261 & 0.242 & 0.017 & 35.058 & 0.268 & geometry / direct-logit & collateral \\
\bottomrule
\end{tabular}%
}
\end{center}
\end{table}
\FloatBarrier

%% file: appendix_heldout_screening_details.tex
\section{Held-out Screening Details}

\begin{figure}[!ht]
\centering
\includegraphics[width=0.95\linewidth,height=0.33\textheight,keepaspectratio]{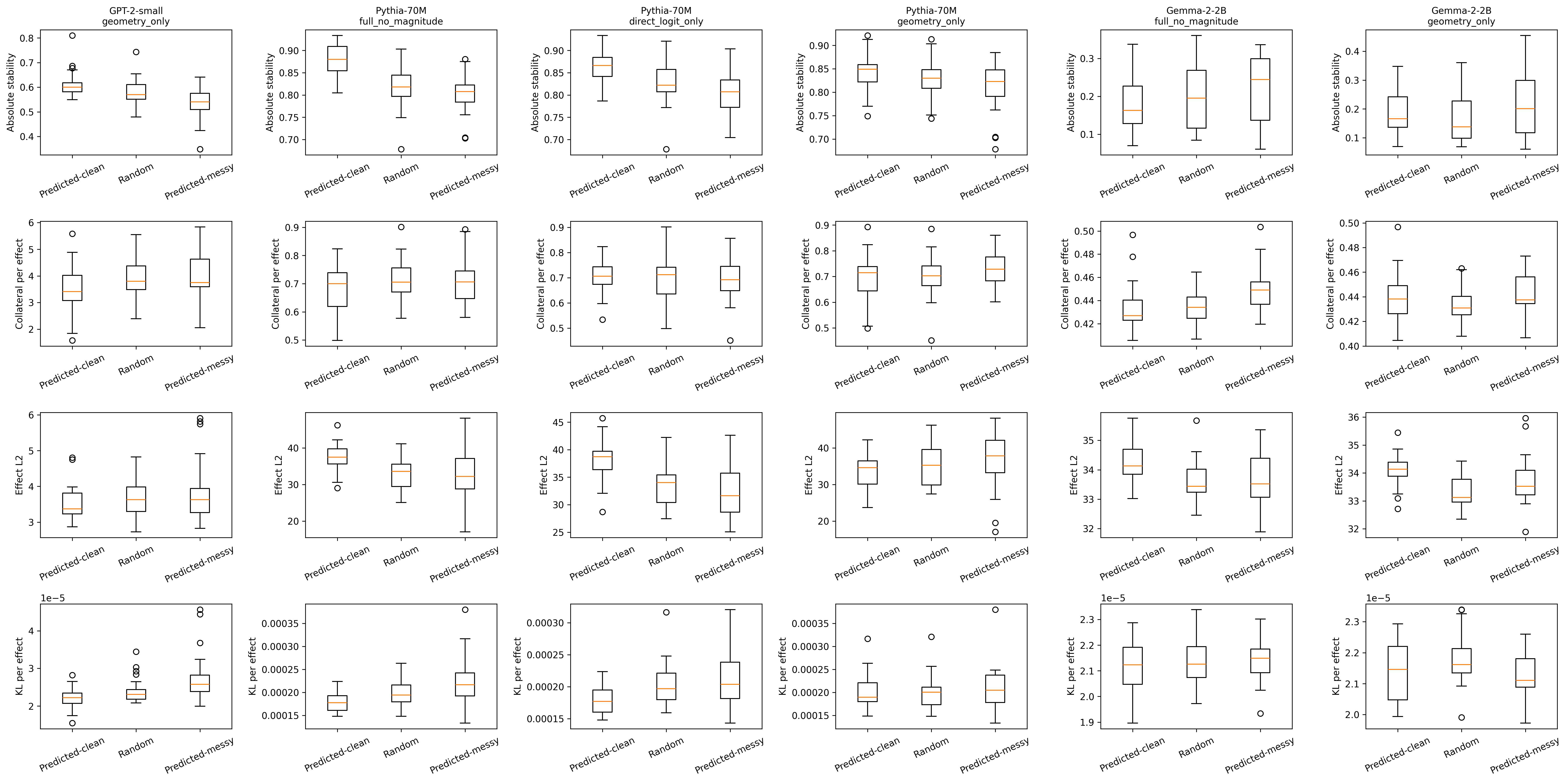}
\caption[Full screening distributions]{Full screening distributions. Distributions of stability, collateral per effect, effect magnitude, and KL per effect for all held-out screening selectors across GPT-2-small, Pythia-70M, and Gemma-2-2B. Corresponding file: \protect\path{figure_D1_full_screening_distributions.png}.}
\label{fig:appendix-full-screening-distributions}

\vspace{8pt}

\includegraphics[width=0.50\linewidth,height=0.32\textheight,keepaspectratio]{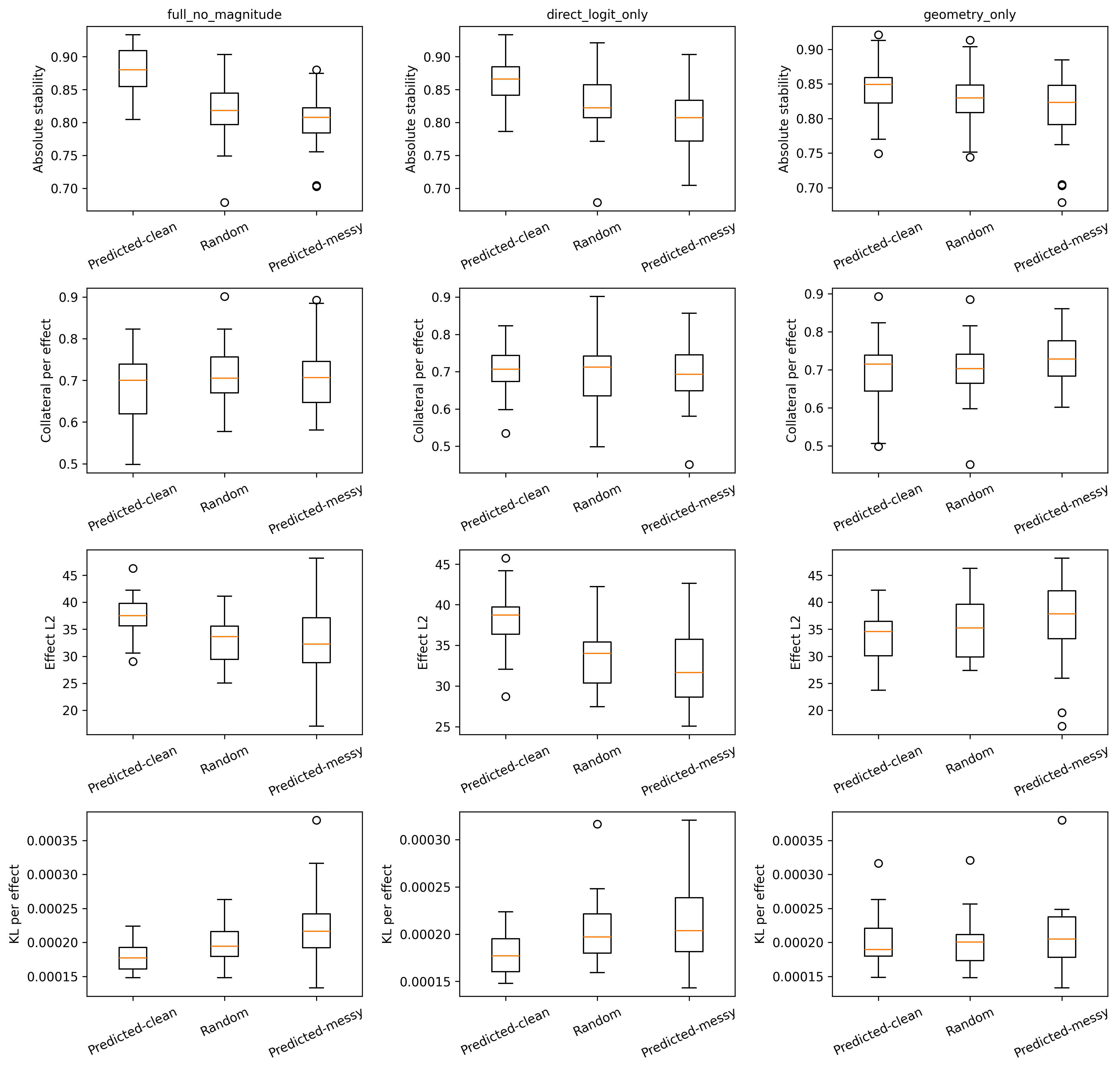}\hspace{0.05\linewidth}%
\includegraphics[width=0.37\linewidth,height=0.32\textheight,keepaspectratio]{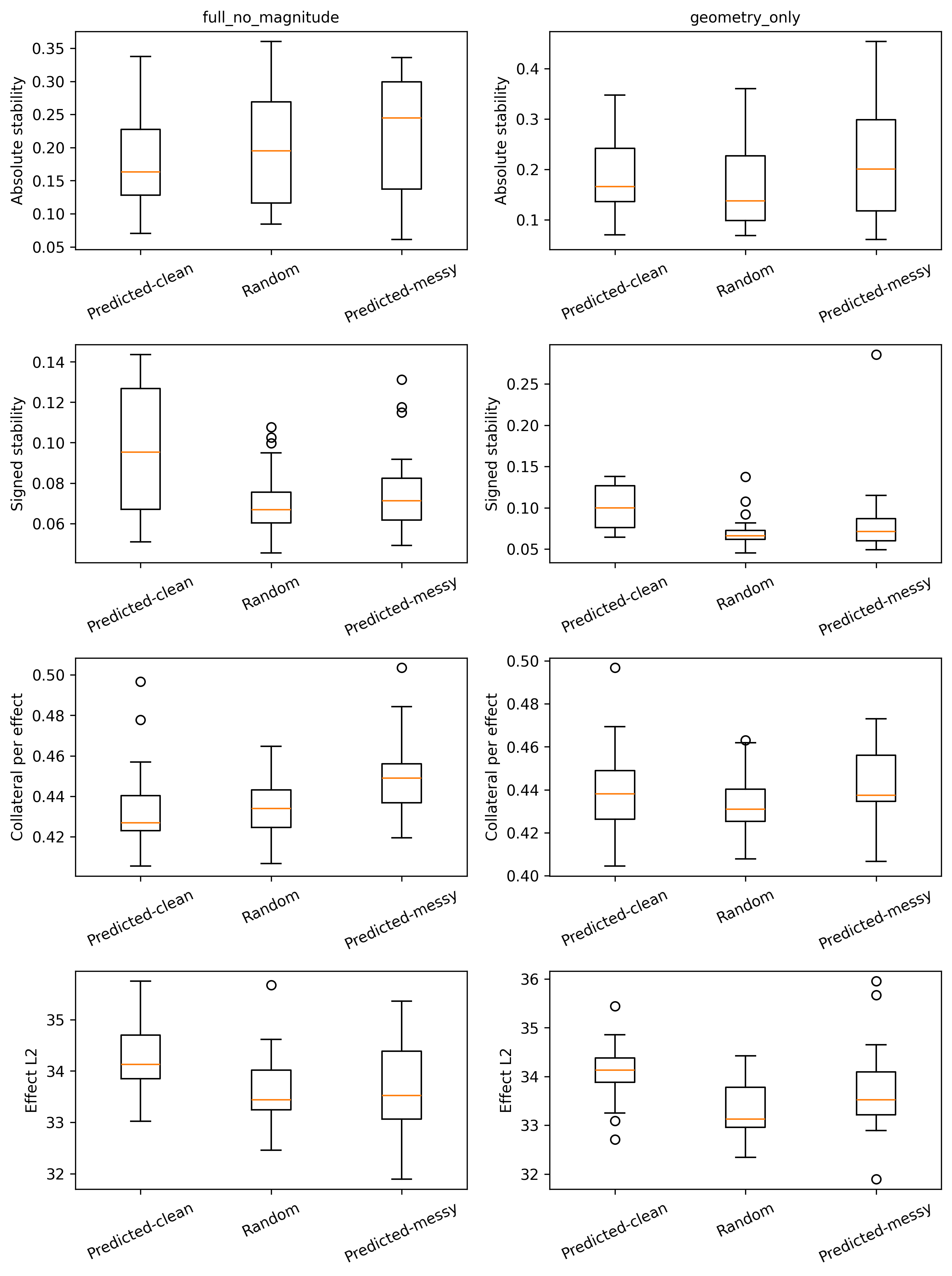}
\caption[Pythia selector comparison]{Pythia selector comparison. Held-out screening distributions for the full-no-magnitude, direct-logit-only, and geometry-only selectors in Pythia-70M. Corresponding file: \protect\path{figure_D2_pythia_selector_comparison.png}.}
\label{fig:appendix-pythia-selector-comparison}
\caption[Gemma partial-transfer diagnostic]{Gemma partial-transfer diagnostic. Held-out screening distributions for the full-no-magnitude and geometry-only selectors in Gemma-2-2B, including absolute stability, signed stability, collateral per effect, and effect magnitude. Corresponding file: \protect\path{figure_D3_gemma_partial_transfer_diagnostic.png}.}
\label{fig:appendix-gemma-partial-transfer}
\end{figure}

This appendix reports the full held-out screening results underlying the main-text screening summary. Table~\ref{tab:appendix-screening-statistical-tests} provides the complete group-level statistical tests for all models, selectors, and metrics. Table~\ref{tab:appendix-screening-feature-ids} lists the feature IDs assigned to predicted-clean, predicted-messy, and random-control groups. Figures~\ref{fig:appendix-full-screening-distributions}--\ref{fig:appendix-gemma-partial-transfer} visualize the full screening distributions, including the model-specific selector comparisons for Pythia-70M and Gemma-2-2B. Section~\ref{app:screening-robustness} reports multiple-comparison and effect-size balance checks for the main clean--messy screening contrasts.

Table~\ref{tab:appendix-screening-statistical-tests} pivots the CSV into one row per model, selector, and metric; the final columns report the three pairwise two-sided Mann--Whitney p-values.

\begin{table}[!ht]
\caption[Full screening statistical tests]{Full screening statistical tests. Complete two-sided Mann--Whitney comparisons among predicted-clean, predicted-messy, and random-control groups for all screening metrics, selectors, and models. Corresponding file: \protect\path{table_D1_full_screening_statistical_tests.csv}.}
\label{tab:appendix-screening-statistical-tests}
\begin{center}
\fontsize{4.8}{4.95}\selectfont
\setlength{\tabcolsep}{1.2pt}
\renewcommand{\arraystretch}{0.70}
\resizebox{\linewidth}{!}{%
\begin{tabular}{@{}lllrrrrrrr@{}}
\hline
\textbf{Model} & \textbf{Selector} & \textbf{Metric} & \textbf{$n$} & \textbf{Clean} & \textbf{Messy} & \textbf{Random} & \textbf{$p_{C,M}$} & \textbf{$p_{C,R}$} & \textbf{$p_{M,R}$}\\
\hline
GPT-2-small & geometry only & Abs. stability & 32 & 0.610 & 0.538 & 0.578 & 4.0e-06 & 0.004 & 0.012\\
GPT-2-small & geometry only & Signed stability & 32 & 0.610 & 0.538 & 0.578 & 3.7e-06 & 0.004 & 0.011\\
GPT-2-small & geometry only & Effect $L_2$ & 32 & 3.52 & 3.80 & 3.64 & 0.118 & 0.157 & 0.783\\
GPT-2-small & geometry only & Effect CV & 32 & 0.298 & 0.323 & 0.311 & 0.007 & 0.672 & 0.097\\
GPT-2-small & geometry only & Downstream feature $L_2$ & 32 & 0.331 & 0.417 & 0.369 & 8.1e-04 & 0.038 & 0.100\\
GPT-2-small & geometry only & Downstream $L_2$/effect & 32 & 0.095 & 0.110 & 0.102 & 0.001 & 0.095 & 0.089\\
GPT-2-small & geometry only & Downstream count $>0.05$ & 32 & 12.1 & 15.1 & 13.8 & 0.005 & 0.063 & 0.262\\
GPT-2-small & geometry only & Downstream count/effect & 32 & 3.49 & 4.01 & 3.83 & 0.020 & 0.082 & 0.394\\
GPT-2-small & geometry only & Effective moved downstream & 32 & 23.3 & 23.4 & 23.5 & 0.825 & 0.268 & 0.409\\
GPT-2-small & geometry only & KL clean to steered & 32 & 0.000 & 0.000 & 0.000 & 1.7e-04 & 0.010 & 0.063\\
GPT-2-small & geometry only & KL/effect & 32 & 0.000 & 0.000 & 0.000 & 1.1e-06 & 0.031 & 9.3e-04\\
Pythia-70M & full no-mag. & Abs. stability & 30 & 0.879 & 0.803 & 0.816 & 1.0e-08 & 1.2e-07 & 0.162\\
Pythia-70M & full no-mag. & Signed stability & 30 & 0.879 & 0.803 & 0.816 & 1.0e-08 & 1.2e-07 & 0.162\\
Pythia-70M & full no-mag. & Effect $L_2$ & 30 & 37.6 & 32.9 & 33.2 & 0.002 & 9.8e-05 & 0.900\\
Pythia-70M & full no-mag. & Effect CV & 30 & 0.260 & 0.264 & 0.269 & 0.404 & 0.217 & 0.579\\
Pythia-70M & full no-mag. & Downstream feature $L_2$ & 30 & 0.581 & 0.549 & 0.561 & 0.429 & 0.631 & 0.620\\
Pythia-70M & full no-mag. & Downstream $L_2$/effect & 30 & 0.015 & 0.017 & 0.017 & 0.008 & 0.011 & 0.784\\
Pythia-70M & full no-mag. & Downstream count $>0.05$ & 30 & 25.9 & 23.3 & 23.7 & 0.086 & 0.030 & 0.706\\
Pythia-70M & full no-mag. & Downstream count/effect & 30 & 0.682 & 0.707 & 0.715 & 0.620 & 0.284 & 0.610\\
Pythia-70M & full no-mag. & Effective moved downstream & 30 & 56.9 & 56.1 & 55.8 & 0.022 & 0.050 & 0.912\\
Pythia-70M & full no-mag. & KL clean to steered & 30 & 0.007 & 0.007 & 0.006 & 0.066 & 0.971 & 0.018\\
Pythia-70M & full no-mag. & KL/effect & 30 & 0.000 & 0.000 & 0.000 & 2.6e-05 & 0.004 & 0.017\\
Pythia-70M & direct logit only & Abs. stability & 30 & 0.864 & 0.807 & 0.829 & 8.9e-06 & 0.004 & 0.029\\
Pythia-70M & direct logit only & Signed stability & 30 & 0.864 & 0.807 & 0.829 & 8.9e-06 & 0.004 & 0.029\\
Pythia-70M & direct logit only & Effect $L_2$ & 30 & 38.0 & 32.6 & 33.6 & 1.5e-04 & 1.7e-05 & 0.240\\
Pythia-70M & direct logit only & Effect CV & 30 & 0.242 & 0.289 & 0.268 & 3.6e-04 & 0.022 & 0.115\\
Pythia-70M & direct logit only & Downstream feature $L_2$ & 30 & 0.575 & 0.559 & 0.548 & 0.318 & 0.122 & 0.971\\
Pythia-70M & direct logit only & Downstream $L_2$/effect & 30 & 0.015 & 0.017 & 0.016 & 0.002 & 0.052 & 0.355\\
Pythia-70M & direct logit only & Downstream count $>0.05$ & 30 & 26.9 & 22.6 & 23.3 & 3.9e-04 & 0.001 & 0.492\\
Pythia-70M & direct logit only & Downstream count/effect & 30 & 0.708 & 0.694 & 0.694 & 0.420 & 0.728 & 0.877\\
Pythia-70M & direct logit only & Effective moved downstream & 30 & 57.6 & 55.3 & 56.0 & 0.003 & 0.030 & 0.212\\
Pythia-70M & direct logit only & KL clean to steered & 30 & 0.007 & 0.007 & 0.007 & 0.923 & 0.830 & 0.971\\
Pythia-70M & direct logit only & KL/effect & 30 & 0.000 & 0.000 & 0.000 & 0.002 & 0.001 & 0.492\\
Pythia-70M & geometry only & Abs. stability & 30 & 0.843 & 0.815 & 0.830 & 0.028 & 0.154 & 0.404\\
Pythia-70M & geometry only & Signed stability & 30 & 0.843 & 0.815 & 0.830 & 0.028 & 0.154 & 0.404\\
Pythia-70M & geometry only & Effect $L_2$ & 30 & 33.7 & 36.3 & 34.9 & 0.056 & 0.501 & 0.206\\
Pythia-70M & geometry only & Effect CV & 30 & 0.271 & 0.250 & 0.266 & 0.028 & 0.379 & 0.196\\
Pythia-70M & geometry only & Downstream feature $L_2$ & 30 & 0.582 & 0.612 & 0.539 & 0.311 & 0.318 & 0.005\\
Pythia-70M & geometry only & Downstream $L_2$/effect & 30 & 0.017 & 0.017 & 0.015 & 0.900 & 0.011 & 0.006\\
Pythia-70M & geometry only & Downstream count $>0.05$ & 30 & 23.6 & 26.6 & 24.4 & 0.029 & 0.589 & 0.039\\
Pythia-70M & geometry only & Downstream count/effect & 30 & 0.698 & 0.733 & 0.699 & 0.122 & 1.000 & 0.115\\
Pythia-70M & geometry only & Effective moved downstream & 30 & 55.0 & 55.7 & 57.4 & 1.000 & 0.021 & 0.003\\
Pythia-70M & geometry only & KL clean to steered & 30 & 0.007 & 0.007 & 0.007 & 0.018 & 0.853 & 0.080\\
Pythia-70M & geometry only & KL/effect & 30 & 0.000 & 0.000 & 0.000 & 0.501 & 0.959 & 0.429\\
Gemma-2-2B & full no-mag. & Abs. stability & 25 & 0.174 & 0.214 & 0.201 & 0.125 & 0.426 & 0.712\\
Gemma-2-2B & full no-mag. & Signed stability & 25 & 0.096 & 0.077 & 0.071 & 0.050 & 0.006 & 0.352\\
Gemma-2-2B & full no-mag. & Effect $L_2$ & 25 & 34.2 & 33.7 & 33.6 & 0.044 & 0.010 & 0.786\\
Gemma-2-2B & full no-mag. & Effect CV & 25 & 0.453 & 0.431 & 0.446 & 0.461 & 0.938 & 0.727\\
Gemma-2-2B & full no-mag. & Downstream feature $L_2$ & 25 & 1.87 & 1.90 & 1.84 & 0.535 & 0.277 & 0.068\\
Gemma-2-2B & full no-mag. & Downstream $L_2$/effect & 25 & 0.055 & 0.056 & 0.055 & 0.099 & 0.985 & 0.125\\
Gemma-2-2B & full no-mag. & Downstream count $>0.05$ & 25 & 14.8 & 15.1 & 14.6 & 0.041 & 0.410 & 0.007\\
Gemma-2-2B & full no-mag. & Downstream count/effect & 25 & 0.433 & 0.450 & 0.434 & 0.002 & 0.415 & 0.003\\
Gemma-2-2B & full no-mag. & Effective moved downstream & 25 & 18.5 & 18.6 & 18.7 & 0.786 & 0.252 & 0.497\\
Gemma-2-2B & full no-mag. & KL clean to steered & 25 & 0.001 & 0.001 & 0.001 & 0.641 & 0.614 & 0.831\\
Gemma-2-2B & full no-mag. & KL/effect & 25 & 0.000 & 0.000 & 0.000 & 0.362 & 0.600 & 0.892\\
Gemma-2-2B & geometry only & Abs. stability & 25 & 0.196 & 0.212 & 0.170 & 0.846 & 0.174 & 0.181\\
Gemma-2-2B & geometry only & Signed stability & 25 & 0.101 & 0.082 & 0.071 & 7.4e-04 & 3.0e-05 & 0.404\\
Gemma-2-2B & geometry only & Effect $L_2$ & 25 & 34.1 & 33.8 & 33.3 & 0.040 & 2.1e-04 & 0.040\\
Gemma-2-2B & geometry only & Effect CV & 25 & 0.451 & 0.450 & 0.427 & 0.727 & 0.304 & 0.561\\
Gemma-2-2B & geometry only & Downstream feature $L_2$ & 25 & 1.89 & 1.87 & 1.84 & 0.252 & 0.052 & 0.322\\
Gemma-2-2B & geometry only & Downstream $L_2$/effect & 25 & 0.055 & 0.055 & 0.055 & 0.509 & 0.415 & 0.892\\
Gemma-2-2B & geometry only & Downstream count $>0.05$ & 25 & 15.0 & 15.0 & 14.4 & 0.992 & 5.5e-04 & 9.7e-04\\
Gemma-2-2B & geometry only & Downstream count/effect & 25 & 0.440 & 0.444 & 0.434 & 0.269 & 0.277 & 0.017\\
Gemma-2-2B & geometry only & Effective moved downstream & 25 & 18.4 & 18.6 & 18.7 & 0.130 & 0.071 & 0.614\\
Gemma-2-2B & geometry only & KL clean to steered & 25 & 0.001 & 0.001 & 0.001 & 0.415 & 0.892 & 0.372\\
Gemma-2-2B & geometry only & KL/effect & 25 & 0.000 & 0.000 & 0.000 & 0.907 & 0.214 & 0.060\\
\hline
\end{tabular}%
}
\end{center}
\end{table}

Table~\ref{tab:appendix-screening-feature-ids} groups the selected feature IDs by model, selector, and screening group while preserving the IDs listed in the source CSV.

\begin{table}[!ht]
\caption[Selected screening feature IDs]{Selected screening feature IDs. Feature IDs selected into predicted-clean, predicted-messy, and random-control groups for each model and selector. Corresponding file: \protect\path{table_D2_selected_screening_feature_ids.csv}.}
\label{tab:appendix-screening-feature-ids}
\begin{center}
\fontsize{6.15}{6.35}\selectfont
\setlength{\tabcolsep}{1pt}
\renewcommand{\arraystretch}{0.94}
\begin{tabular}{@{}>{\raggedright\arraybackslash}p{0.14\linewidth}>{\raggedright\arraybackslash}p{0.17\linewidth}>{\raggedright\arraybackslash}p{0.15\linewidth}r>{\raggedright\arraybackslash}p{0.43\linewidth}@{}}
\hline
\textbf{Model} & \textbf{Selector} & \textbf{Group} & \textbf{$n$} & \textbf{Feature IDs}\\
\hline
GPT-2-small & geometry only & predicted clean & 32 & 19163, 10918, 21830, 10152, 18029, 16920, 24019, 4419, 10201, 2497, 14054, 13851, 14220, 22672, 4838, 12593, 542, 20602, 10462, 13424, 14211, 5215, 18168, 17907, 14393, 8081, 16947, 12610, 3699, 20839, 23231, 15946\\
GPT-2-small & geometry only & predicted messy & 32 & 920, 5672, 7502, 19190, 6219, 8466, 3657, 1680, 20395, 9580, 23564, 9278, 9626, 20929, 18053, 12409, 22509, 14440, 1351, 1128, 5291, 3204, 4965, 22860, 12100, 13660, 2153, 17680, 4434, 788, 1069, 17054\\
GPT-2-small & geometry only & random control & 32 & 6798, 9322, 7205, 14123, 10104, 13751, 20062, 20749, 21222, 16926, 6293, 12885, 11041, 23868, 6953, 11844, 21529, 23441, 8822, 14714, 18525, 8928, 11951, 21877, 11722, 23032, 9887, 21746, 15963, 7795, 22792, 2360\\
Pythia-70M & full no-mag. & predicted clean & 30 & 28480, 24554, 23153, 3580, 4823, 2024, 32640, 17806, 5857, 30702, 776, 7817, 10569, 11911, 31683, 24286, 5708, 15639, 11720, 13148, 21344, 15599, 7834, 4811, 20047, 13285, 19976, 9914, 6621, 24899\\
Pythia-70M & full no-mag. & predicted messy & 30 & 6798, 15069, 31878, 17938, 1618, 23156, 22612, 21980, 19907, 32299, 27729, 31458, 23430, 6034, 13121, 6307, 6932, 24025, 31360, 10492, 11570, 26641, 16353, 9150, 31151, 16899, 23687, 17148, 465, 26638\\
Pythia-70M & full no-mag. & random control & 30 & 16642, 21620, 14459, 3120, 12802, 29601, 8024, 1792, 26041, 17653, 20028, 3866, 22031, 26795, 6382, 22990, 8865, 18837, 20241, 5105, 22024, 9363, 18145, 8423, 7277, 29500, 23074, 16434, 11654, 16052\\
Pythia-70M & direct logit only & predicted clean & 30 & 23153, 5708, 32465, 2024, 4823, 28480, 19976, 3120, 3580, 11911, 23507, 21344, 23425, 17806, 20047, 11720, 5857, 24554, 13148, 13285, 24349, 14643, 7817, 32640, 776, 23074, 29500, 25457, 8865, 4811\\
Pythia-70M & direct logit only & predicted messy & 30 & 15069, 14483, 22024, 1988, 6932, 1618, 14050, 23156, 24025, 12295, 17938, 26041, 22031, 3866, 6034, 11570, 29601, 13121, 32299, 16353, 27729, 19907, 23687, 10492, 31222, 20028, 19200, 8024, 23430, 31151\\
Pythia-70M & direct logit only & random control & 30 & 15599, 21620, 26795, 12802, 21980, 1792, 27133, 18145, 7834, 20241, 31458, 17148, 23214, 5105, 30702, 22990, 6382, 24286, 465, 16642, 16434, 9150, 7277, 6307, 31683, 17653, 16052, 6621, 11654, 8423\\
Pythia-70M & geometry only & predicted clean & 30 & 15599, 3587, 6382, 22297, 30702, 29601, 17653, 4811, 31151, 12802, 28480, 18837, 9150, 24899, 20028, 24286, 6621, 5857, 15639, 5708, 776, 5105, 12295, 30974, 6307, 1156, 18145, 16899, 16353, 14050\\
Pythia-70M & geometry only & predicted messy & 30 & 6798, 29730, 10492, 24349, 21980, 27729, 22612, 23425, 31878, 6034, 17148, 9185, 17938, 21620, 11911, 31458, 27133, 19334, 13285, 23214, 2054, 23687, 15069, 3866, 13121, 23507, 465, 21344, 6932, 23156\\
Pythia-70M & geometry only & random control & 30 & 20047, 1618, 1988, 14483, 8024, 26641, 16860, 31683, 13148, 24025, 32465, 22031, 19976, 16642, 2024, 14643, 11720, 18196, 8423, 9914, 31222, 627, 16557, 26871, 3580, 23074, 19200, 31360, 26041, 4823\\
Gemma-2-2B & full no-mag. & predicted clean & 25 & 13722, 11577, 3544, 1233, 2164, 6939, 1874, 6283, 15715, 3603, 11922, 1389, 12074, 12676, 14024, 8780, 11238, 12704, 4634, 13016, 5176, 5693, 14664, 5220, 1218\\
Gemma-2-2B & full no-mag. & predicted messy & 25 & 1257, 15281, 250, 9939, 8260, 3582, 6600, 10202, 7970, 10009, 12012, 3210, 5639, 11521, 16343, 12963, 361, 6384, 4636, 2611, 13735, 11483, 9066, 8978, 8904\\
Gemma-2-2B & full no-mag. & random control & 25 & 1401, 14833, 11539, 15605, 12770, 6817, 12468, 12105, 347, 7698, 2475, 15616, 6737, 5839, 2235, 8267, 14291, 9517, 3284, 15611, 7816, 4245, 12333, 12060, 15670\\
Gemma-2-2B & geometry only & predicted clean & 25 & 12770, 11577, 6939, 8780, 11238, 14527, 3213, 6283, 2164, 14664, 13722, 1233, 1759, 9259, 10009, 1218, 12676, 3675, 15670, 12704, 9346, 10202, 2475, 3284, 7255\\
Gemma-2-2B & geometry only & predicted messy & 25 & 3019, 8146, 7970, 13646, 8978, 5078, 1257, 8949, 9939, 3582, 250, 9016, 2235, 1401, 6737, 13735, 15611, 8260, 5693, 15281, 11483, 6384, 3210, 2579, 9066\\
Gemma-2-2B & geometry only & random control & 25 & 13829, 12074, 11539, 2201, 5220, 11988, 7816, 14024, 4143, 7698, 11043, 4636, 5487, 6815, 15605, 5839, 14833, 931, 16109, 13959, 11521, 8904, 12333, 11922, 7873\\
\hline
\end{tabular}
\end{center}
\end{table}

\FloatBarrier

\FloatBarrier

\subsection{Screening robustness checks}
\label{app:screening-robustness}

This subsection reports additional robustness checks for the held-out screening evaluation. First, because the screening analysis reports multiple selector-by-axis contrasts, we provide Holm and Benjamini--Hochberg adjusted p-values for both the full set of reported contrasts and the pre-specified main-selector contrasts. Second, because differences in realized intervention magnitude could confound clean--messy comparisons, we report post-selection effect-size balance between predicted-clean and predicted-messy groups. These checks do not define new screening criteria; they are included to clarify which held-out effects are robust to multiplicity and which should be interpreted cautiously because of effect-magnitude imbalance.

\begin{table}[!htbp]
\caption{Multiple-comparison correction for held-out screening contrasts. Raw p-values are two-sided Mann--Whitney tests comparing predicted-clean and predicted-messy groups. Adjusted p-values are computed using Holm and Benjamini--Hochberg correction both over all reported selector-by-axis contrasts and over the pre-specified main-selector contrasts. The strongest screening conclusions remain after correction, while marginal auxiliary selector effects should be interpreted cautiously.}
\label{tab:screening-multiple-corrections}
\begin{center}
\tiny
\setlength{\tabcolsep}{1pt}
\renewcommand{\arraystretch}{0.86}
\resizebox{\linewidth}{!}{%
\begin{tabular}{@{}lllrrrrrrrrlllll@{}}
\toprule
Model & Selector & Metric & Clean & Messy & $\Delta$ & $p$ raw & Holm all & BH all & Holm pre & BH pre & Main & Holm all sig. & BH all sig. & Holm pre sig. & BH pre sig. \\
\midrule
GPT-2-small & geometry only & stability & 0.610 & 0.538 & 0.071 & 3.98e-06 & 6.77e-05 & 3.59e-05 & 2.79e-05 & 1.59e-05 & True & True & True & True & True \\
GPT-2-small & geometry only & collateral/effect & 3.486 & 4.010 & -0.523 & 0.020 & 0.258 & 0.059 & 0.079 & 0.032 & True & False & False & False & True \\
Pythia-70M & direct logit only & stability & 0.864 & 0.807 & 0.057 & 8.88e-06 & 1.421e-04 & 5.33e-05 & -- & -- & False & True & True & False & False \\
Pythia-70M & direct logit only & collateral/effect & 0.708 & 0.694 & 0.014 & 0.420 & 1.000 & 0.473 & -- & -- & False & False & False & False & False \\
Pythia-70M & full no-mag. & stability & 0.879 & 0.803 & 0.076 & 1.01e-08 & 1.82e-07 & 1.82e-07 & 8.08e-08 & 8.08e-08 & True & True & True & True & True \\
Pythia-70M & full no-mag. & collateral/effect & 0.682 & 0.707 & -0.025 & 0.620 & 1.000 & 0.657 & 0.620 & 0.620 & True & False & False & False & False \\
Pythia-70M & geometry only & stability & 0.843 & 0.815 & 0.029 & 0.028 & 0.338 & 0.072 & -- & -- & False & False & False & False & False \\
Pythia-70M & geometry only & collateral/effect & 0.698 & 0.733 & -0.035 & 0.122 & 1.000 & 0.205 & -- & -- & False & False & False & False & False \\
Gemma-2-2B & full no-mag. & stability & 0.174 & 0.214 & -0.040 & 0.125 & 1.000 & 0.205 & 0.376 & 0.167 & True & False & False & False & False \\
Gemma-2-2B & full no-mag. & collateral/effect & 0.433 & 0.450 & -0.017 & 0.002 & 0.025 & 0.006 & 0.009 & 0.004 & True & True & True & True & True \\
Gemma-2-2B & geometry only & stability & 0.196 & 0.212 & -0.016 & 0.846 & 1.000 & 0.846 & -- & -- & False & False & False & False & False \\
Gemma-2-2B & geometry only & collateral/effect & 0.440 & 0.444 & -0.004 & 0.269 & 1.000 & 0.384 & -- & -- & False & False & False & False & False \\
Llama-3.1-8B & direct logit only & stability & 0.219 & 0.205 & 0.015 & 0.068 & 0.682 & 0.136 & -- & -- & False & False & False & False & False \\
Llama-3.1-8B & direct logit only & collateral/effect & 0.236 & 0.256 & -0.020 & 0.332 & 1.000 & 0.398 & -- & -- & False & False & False & False & False \\
Llama-3.1-8B & full no-mag. & stability & 0.220 & 0.212 & 0.009 & 0.277 & 1.000 & 0.384 & 0.554 & 0.317 & True & False & False & False & False \\
Llama-3.1-8B & full no-mag. & collateral/effect & 0.209 & 0.281 & -0.072 & 1.322e-04 & 0.002 & 5.948e-04 & 7.931e-04 & 3.525e-04 & True & True & True & True & True \\
Llama-3.1-8B & geometry only & stability & 0.216 & 0.209 & 0.007 & 0.332 & 1.000 & 0.398 & -- & -- & False & False & False & False & False \\
Llama-3.1-8B & geometry only & collateral/effect & 0.221 & 0.267 & -0.046 & 0.040 & 0.437 & 0.089 & -- & -- & False & False & False & False & False \\
\bottomrule
\end{tabular}%
}
\end{center}
\end{table}

\begin{table}[!htbp]
\caption{Effect-size balance between predicted-clean and predicted-messy groups in held-out screening. Effect magnitude is measured by realized logit-effect $L_2$ on fresh contexts. Non-significant differences indicate that the clean--messy contrast is not explained by realized intervention magnitude alone; significant differences identify settings where screening results should be interpreted with additional caution.}
\label{tab:effect-size-balance}
\begin{center}
\small
\setlength{\tabcolsep}{3pt}
\resizebox{\linewidth}{!}{%
\begin{tabular}{@{}llrrrrll@{}}
\toprule
Model & Selector & Clean effect $L_2$ & Messy effect $L_2$ & $\Delta$ & $p$ effect $L_2$ & Balanced at 0.05 & Main selector \\
\midrule
GPT-2-small & geometry only & 3.516 & 3.803 & -0.287 & 0.118 & True & True \\
Pythia-70M & direct logit only & 37.995 & 32.550 & 5.445 & 1.493e-04 & False & False \\
Pythia-70M & full no-mag. & 37.622 & 32.876 & 4.746 & 0.002 & False & True \\
Pythia-70M & geometry only & 33.750 & 36.302 & -2.552 & 0.056 & True & False \\
Gemma-2-2B & full no-mag. & 34.198 & 33.678 & 0.521 & 0.044 & False & True \\
Gemma-2-2B & geometry only & 34.099 & 33.775 & 0.324 & 0.040 & False & False \\
Llama-3.1-8B & direct logit only & 13.063 & 12.618 & 0.444 & 0.698 & True & False \\
Llama-3.1-8B & full no-mag. & 12.815 & 12.845 & -0.030 & 0.742 & True & True \\
Llama-3.1-8B & geometry only & 12.956 & 12.759 & 0.197 & 0.485 & True & False \\
\bottomrule
\end{tabular}%
}
\end{center}
\end{table}

\begin{table}[!htbp]
\caption{GPT-2-small intervention-strength sweep for held-out screening. The same predicted-clean, random-control, and predicted-messy feature groups are re-evaluated under additive steering strengths $\alpha \in \{0.5,1.0,2.0\}$. Stability remains higher for predicted-clean features at all three strengths, while collateral per effect is significantly lower at $\alpha=0.5$ and $\alpha=1.0$ but attenuates at $\alpha=2.0$. Effect $L_2$ remains balanced at all strengths.}
\label{tab:gpt2-alpha-sweep}
\begin{center}
\small
\setlength{\tabcolsep}{4pt}
\resizebox{\linewidth}{!}{%
\begin{tabular}{@{}rrrrrrrl@{}}
\toprule
$\alpha$ & Stability clean & Stability messy & $p$ & Collateral/effect clean & Collateral/effect messy & $p$ & Effect $L_2$ balanced? \\
\midrule
0.5 & 0.6029 & 0.5288 & $7.5\times10^{-7}$ & 0.7638 & 1.3525 & $1.58\times10^{-5}$ & Yes \\
1.0 & 0.6031 & 0.5291 & $8.04\times10^{-7}$ & 1.7115 & 2.0987 & $5.73\times10^{-4}$ & Yes \\
2.0 & 0.6036 & 0.5300 & $9.22\times10^{-7}$ & 1.5551 & 1.6392 & 0.200 & Yes \\
\bottomrule
\end{tabular}%
}
\end{center}
\end{table}

%% file: appendix_additional_metric_predictor_definitions.tex
\section{Additional Metric and Predictor Definitions}
\label{app:additional-metric-predictor-definitions}

This appendix provides the detailed metric and predictor equations omitted from the main Methodology section for readability.

\subsection{Auxiliary steering labels}

In addition to the primary stability and collateral labels, we compute auxiliary effect-magnitude and distribution-shift metrics. The coefficient of variation of the logit-effect magnitude is
\[
\mathrm{CV}_f =
\frac{
\operatorname{Std}_{x_i \in \mathcal{C}_f}
\left(
\left\|\Delta \ell_i^{(f)}\right\|_2
\right)
}{
\operatorname{Mean}_{x_i \in \mathcal{C}_f}
\left(
\left\|\Delta \ell_i^{(f)}\right\|_2
\right)
+
\varepsilon
}.
\]

We also compute the KL divergence between the clean and steered next-token distributions. Let
\[
p_i = \mathrm{softmax}(\ell_i),
\qquad
q_i^{(f)} = \mathrm{softmax}(\ell_i^{(f)}).
\]
The mean KL shift is
\[
K_f =
\frac{1}{|\mathcal{C}_f|}
\sum_{x_i \in \mathcal{C}_f}
D_{\mathrm{KL}}
\left(
p_i \,\|\, q_i^{(f)}
\right),
\]
and the normalized KL-per-effect score is
\[
\widetilde{K}_f =
\frac{K_f}{E_f + \varepsilon}.
\]

\subsection{Decoder geometry predictors}

Let $d_f$ be the decoder vector for feature $f$ and $e_f$ its corresponding encoder vector. We compute decoder norm, encoder norm, and encoder-decoder alignment:
\[
\|d_f\|_2,
\qquad
\|e_f\|_2,
\qquad
A_f = \cos(e_f, d_f).
\]

Decoder crowding measures how close $d_f$ is to neighboring decoder directions. Let $\mathcal{N}_K(f)$ be the set of the $K$ decoder vectors with largest absolute cosine similarity to $d_f$, excluding $f$. We define
\[
G_f^{\mathrm{mean}}
=
\frac{1}{K}
\sum_{g \in \mathcal{N}_K(f)}
\left|\cos(d_f, d_g)\right|,
\]
and
\[
G_f^{\mathrm{max}}
=
\max_{g \ne f}
\left|\cos(d_f, d_g)\right|.
\]

\subsection{Activation statistics}

Let $a_f(x_i)$ be the activation of feature $f$ on context $x_i$, and let $p_f = \mathrm{freq}(f)$. We compute activation moments such as mean, standard deviation, maximum activation, and kurtosis. We also compute binary firing entropy:
\[
H_f^{\mathrm{bin}}
=
-p_f\log(p_f + \varepsilon)
-(1 - p_f)\log(1 - p_f + \varepsilon).
\]

For activation entropy, define
\[
r_{f,i}
=
\frac{a_f(x_i)}
{\sum_k a_f(x_k) + \varepsilon}.
\]
The normalized activation entropy is
\[
H_f^{\mathrm{act}}
=
-
\frac{
\sum_i r_{f,i}\log(r_{f,i} + \varepsilon)
}
{\log N}.
\]

\subsection{Co-activation statistics}

Let
\[
B_{i,j} = \mathbf{1}[a_j(x_i) > \epsilon_{\mathrm{fire}}]
\]
indicate whether feature $j$ fires in context $i$. Conditional on feature $f$ firing, the co-activation rate of feature $j$ is
\[
q_{f,j}
=
\frac{
\sum_i B_{i,f}B_{i,j}
}{
\sum_i B_{i,f} + \varepsilon
}.
\]
After normalizing these rates,
\[
\pi_{f,j}
=
\frac{q_{f,j}}
{\sum_k q_{f,k} + \varepsilon},
\]
we compute co-activation entropy:
\[
H_f^{\mathrm{coact}}
=
-
\sum_j
\pi_{f,j}
\log(\pi_{f,j} + \varepsilon).
\]

\subsection{Direct-logit statistics}

Let $W_U$ be the unembedding matrix. The direct-logit vector for feature $f$ is
\[
r_f = d_f W_U.
\]
We compute its norm statistics,
\[
\|r_f\|_2,
\qquad
\|r_f\|_\infty,
\]
and its normalized absolute-mass distribution
\[
s_{f,v}
=
\frac{|r_{f,v}|}
{\sum_{v'} |r_{f,v'}| + \varepsilon}.
\]
The direct-logit entropy is
\[
H_f^{\mathrm{logit}}
=
-
\sum_v
s_{f,v}
\log(s_{f,v} + \varepsilon),
\]
and the top-$k$ absolute-mass fraction is
\[
M_{f,k}
=
\sum_{v \in \operatorname{TopK}(|r_f|; k)}
s_{f,v}.
\]

For Gemma-2-2B, the model's final logit softcap is preserved during forward evaluation. Direct-logit predictors are therefore treated as pre-softcap shape descriptors and are not compared by absolute scale across model families.

%% file: appendix_llama_replication.tex
\section{Llama-3.1-8B Replication}
\label{app:llama}

After the initial three-model evaluation, we added a fourth replication on Llama-3.1-8B to test whether the pre-intervention screening framework extends to a larger and more recent open-weight model. This replication uses Llama Scope TopK residual-stream SAEs and follows the same overall Phase 1--3 predictive pipeline and Phase 4 held-out screening protocol as the main experiments. We include the Llama results in the appendix rather than the main text to keep the primary narrative focused, but the replication strengthens the cross-model evidence: Llama-3.1-8B shows strong predictive performance and a held-out collateral-screening effect, while also reinforcing the paper's conclusion that the transferred axis of modularity is model-dependent.

\begin{table}[!htbp]
\caption{Llama-3.1-8B replication configuration. The replication uses Llama Scope residual-stream SAEs at a mid-depth primary site and a later downstream collateral-measurement site.}
\label{tab:llama-config}
\begin{center}
\small
\begin{tabular}{ll}
\hline
\textbf{Setting} & \textbf{Value} \\
\hline
Model & Llama-3.1-8B \\
Model checkpoint & \texttt{meta-llama/Llama-3.1-8B} \\
SAE release & \texttt{llama\_scope\_lxr\_8x} \\
Primary SAE & \texttt{l16r\_8x} \\
Primary hook & \texttt{blocks.16.hook\_resid\_post} \\
Downstream SAE & \texttt{l20r\_8x} \\
Downstream hook & \texttt{blocks.20.hook\_resid\_post} \\
Dataset & Wikitext-103, \texttt{train[:2\%]} \\
Distinct texts & 8{,}000 \\
Contexts & 2{,}048 \\
Context length & 48 \\
Features & 300 \\
Contexts per type & 16 \\
Intervention & \texttt{fixed\_global\_add} \\
Additive value & $\alpha = 1.0$ \\
Downstream panel size & 1{,}024 \\
Collateral threshold & 0.05 \\
\hline
\end{tabular}
\end{center}
\end{table}

\begin{table}[!htbp]
\caption{Llama-3.1-8B predictive-evaluation summary. Cross-validated Spearman performance for the main steering-label targets. The full no-magnitude predictor set improves strongly over frequency-only and activation-magnitude-only baselines across stability, effect-magnitude, and collateral targets. Corresponding file: \texttt{robust\_phase3\_baseline\_comparison.csv}.}
\label{tab:llama-predictive-summary}
\begin{center}
\small
\setlength{\tabcolsep}{3.5pt}
\begin{tabular}{lrrrrr}
\hline
\textbf{Target} & \textbf{Freq.} & \textbf{Act. mag.} & \textbf{Geom.} & \textbf{Direct logit} & \textbf{Full no-mag.} \\
\hline
Signed stability & $-0.021$ & $-0.024$ & $0.368$ & $0.334$ & $0.447$ \\
Abs. stability & $-0.057$ & $-0.071$ & $0.291$ & $0.221$ & $0.358$ \\
Effect $L_2$ & $-0.039$ & $0.072$ & $0.732$ & $0.690$ & $0.722$ \\
Effect CV & $-0.079$ & $-0.170$ & $0.428$ & $0.431$ & $0.392$ \\
Downstream count $>0.05$ & $0.018$ & $0.247$ & $0.540$ & $0.511$ & $0.631$ \\
Collateral/effect & $0.161$ & $0.230$ & $0.274$ & $0.319$ & $0.474$ \\
Downstream effective moved & $0.131$ & $0.282$ & $0.491$ & $0.351$ & $0.556$ \\
Downstream $L_2$/effect & $0.181$ & $0.252$ & $0.232$ & $0.050$ & $0.327$ \\
\hline
Mean improvement over frequency & \multicolumn{5}{r}{$+0.451$} \\
Mean improvement over activation magnitude & \multicolumn{5}{r}{$+0.386$} \\
\hline
\end{tabular}
\end{center}
\end{table}

\begin{table}[!htbp]
\caption{Strongest Llama-3.1-8B univariate predictor relationships. The strongest overall univariate relationship is decoder norm predicting effect magnitude, but the side-effect-relevant correlations also show substantial signal for downstream collateral and stability. Corresponding file: \texttt{robust\_phase3\_univariate\_correlations.csv}.}
\label{tab:llama-univariate}
\begin{center}
\small
\setlength{\tabcolsep}{4pt}
\begin{tabular}{llrr}
\hline
\textbf{Predictor} & \textbf{Target} & \textbf{Spearman $\rho$} & \textbf{$p$} \\
\hline
decoder norm & effect $L_2$ & $0.727$ & $1.12 \times 10^{-50}$ \\
direct-logit footprint ($L_2$) & effect $L_2$ & $0.672$ & $1.01 \times 10^{-40}$ \\
direct-logit footprint ($L_2$) & downstream count $>0.05$ & $0.464$ & $1.87 \times 10^{-17}$ \\
decoder norm & downstream count $>0.05$ & $0.414$ & $7.44 \times 10^{-14}$ \\
decoder norm & signed stability & $0.366$ & $6.01 \times 10^{-11}$ \\
direct-logit footprint ($\ell_\infty$) & signed stability & $0.350$ & $4.64 \times 10^{-10}$ \\
\hline
\end{tabular}
\end{center}
\end{table}

\begin{table}[!htbp]
\caption{Llama-3.1-8B residualized-stability robustness. Stability labels are residualized against effect magnitude, intervention value, and natural activation before prediction. The full no-magnitude set retains nontrivial residualized signal, unlike the frequency-only and activation-magnitude-only baselines. Corresponding file: \texttt{robust\_phase3\_residualized\_target\_results.csv}.}
\label{tab:llama-residualized}
\begin{center}
\small
\begin{tabular}{llrr}
\hline
\textbf{Original target} & \textbf{Predictor set} & \textbf{Predictors} & \textbf{CV Spearman} \\
\hline
Signed stability & frequency only & 1 & $-0.109$ \\
Signed stability & activation magnitude only & 4 & $-0.139$ \\
Signed stability & geometry only & 5 & $0.148$ \\
Signed stability & direct-logit only & 5 & $0.065$ \\
Signed stability & coactivation only & 3 & $0.207$ \\
Signed stability & full no-magnitude & 16 & $0.366$ \\
Signed stability & full all & 20 & $0.323$ \\
\hline
Abs. stability & frequency only & 1 & $-0.119$ \\
Abs. stability & activation magnitude only & 4 & $-0.177$ \\
Abs. stability & geometry only & 5 & $0.111$ \\
Abs. stability & direct-logit only & 5 & $-0.028$ \\
Abs. stability & coactivation only & 3 & $0.176$ \\
Abs. stability & full no-magnitude & 16 & $0.290$ \\
Abs. stability & full all & 20 & $0.248$ \\
\hline
\end{tabular}
\end{center}
\end{table}

\begin{table}[!htbp]
\caption{Llama-3.1-8B held-out screening performance. Predicted-clean and predicted-messy groups are selected from held-out features and evaluated on fresh Wikitext validation contexts. Stability is mean absolute cosine to the average effect, where higher is better; collateral is downstream count per unit effect, where lower is better. Corresponding files: \texttt{llama31\_phase4\_screening\_summary.csv}, \texttt{llama31\_phase4\_group\_tests.csv}, and \texttt{llama31\_phase4\_fresh\_steering\_eval.csv}.}
\label{tab:llama-screening}
\begin{center}
\scriptsize
\setlength{\tabcolsep}{3pt}
\resizebox{\linewidth}{!}{%
\begin{tabular}{lrrrrrrrrl}
\hline
\textbf{Selector} & \textbf{$n$/group} & \textbf{Contexts} & \textbf{Clean stab.} & \textbf{Messy stab.} & \textbf{$p$ stab.} & \textbf{Clean coll./eff.} & \textbf{Messy coll./eff.} & \textbf{$p$ coll.} & \textbf{Verdict} \\
\hline
full no-magnitude & 25 & 256 & $0.220$ & $0.212$ & $0.277$ & $0.209$ & $0.281$ & $1.32 \times 10^{-4}$ & partial--collateral only \\
geometry only & 25 & 256 & $0.216$ & $0.209$ & $0.332$ & $0.221$ & $0.267$ & $0.0397$ & partial--collateral only \\
direct-logit only & 25 & 256 & $0.219$ & $0.205$ & $0.068$ & $0.236$ & $0.256$ & $0.332$ & weak \\
\hline
\end{tabular}%
}
\end{center}
\end{table}

The Llama replication supports the main paper's method-level claim but not a universal mechanism-level claim. In predictive evaluation, Llama-3.1-8B shows the largest mean improvement over both frequency-only and activation-magnitude-only baselines among the evaluated model settings. In residualized robustness, the full no-magnitude predictor set retains signal on both signed and absolute stability after controlling for magnitude-related variables. In held-out screening, however, the transferred effect appears on the collateral axis rather than the stability axis: the full no-magnitude selector significantly reduces collateral per effect without a significant difference in effect magnitude, but does not significantly improve stability. Thus, Llama strengthens the evidence that SAE steering side effects are predictable in advance, while reinforcing the conclusion that the specific predictor signature and successful screening axis are model-dependent.

%% file: appendix_llama_width_control.tex
\section{Controlled Llama Scope Width Comparison}
\label{app:llama-width-control}

\begin{table}[!htbp]
\caption{Controlled Llama Scope dictionary-width comparison. Both rows use the same Llama-3.1-8B base model, hook sites, dataset, feature count, context count, and additive intervention; only the Llama Scope residual-stream SAE width changes. Predictive gains report mean five-fold CV Spearman improvement over the frequency-only and activation-magnitude-only baselines across primary targets. The 128K dictionary preserves strong pre-intervention predictability but weakens the held-out screening result, indicating that dictionary granularity affects the practical screening axis.}
\label{tab:llama-width-control}
\begin{center}
\small
\setlength{\tabcolsep}{3pt}
\resizebox{\textwidth}{!}{%
\begin{tabular}{@{}llrrlrl@{}}
\toprule
Setting & SAE IDs & $\Delta$ freq. & $\Delta$ act.-mag. & Best primary predictor & $\rho$ & Phase 4 outcome \\
\midrule
Llama Scope 32K & \texttt{l16r\_8x}/\texttt{l20r\_8x} & $+0.451$ & $+0.386$ & decoder norm $\rightarrow$ effect $L_2$ & $0.727$ & collateral only \\
Llama Scope 128K & \texttt{l16r\_32x}/\texttt{l20r\_32x} & $+0.420$ & $+0.266$ & decoder norm $\rightarrow$ effect $L_2$ & $0.638$ & weak; direct-logit gives stability only \\
\bottomrule
\end{tabular}%
}
\end{center}
\end{table}
\begin{table}[!htbp]
\caption{Held-out Phase 4 clean--messy group tests for the controlled Llama Scope 128K setting. Each row compares predicted-clean and predicted-messy features selected by the indicated predictor set. Higher stability is better, while lower collateral/effect and KL/effect are better. The full no-magnitude selector trends in the expected direction on stability and collateral but does not reach significance on the two pre-specified axes; the direct-logit selector gives a significant stability-only effect.}
\label{tab:llama128k-phase4-group-tests}
\begin{center}
\small
\setlength{\tabcolsep}{4pt}
\resizebox{\textwidth}{!}{%
\begin{tabular}{@{}llrrrr@{}}
\toprule
Selector & Metric & Clean mean & Messy mean & $\Delta$ clean--messy & $p$ \\
\midrule
Full no-mag. & Stability, abs. cos. & 0.260 & 0.222 & +0.038 & 0.081 \\
Full no-mag. & Stability, signed cos. & 0.255 & 0.211 & +0.044 & 0.034 \\
Full no-mag. & Collateral/effect & 0.248 & 0.293 & -0.046 & 0.088 \\
Full no-mag. & Effect $L_2$ & 13.825 & 13.002 & +0.824 & 0.252 \\
Full no-mag. & KL/effect & 0.000052 & 0.000055 & -0.000002 & 0.125 \\
\midrule
Geometry only & Stability, abs. cos. & 0.235 & 0.250 & -0.015 & 0.669 \\
Geometry only & Stability, signed cos. & 0.221 & 0.241 & -0.019 & 0.727 \\
Geometry only & Collateral/effect & 0.270 & 0.254 & +0.016 & 0.404 \\
Geometry only & Effect $L_2$ & 13.056 & 13.404 & -0.348 & 0.655 \\
Geometry only & KL/effect & 0.000053 & 0.000053 & +0.000000 & 0.892 \\
\midrule
Direct-logit only & Stability, abs. cos. & 0.258 & 0.210 & +0.048 & 0.012 \\
Direct-logit only & Stability, signed cos. & 0.252 & 0.203 & +0.049 & 0.014 \\
Direct-logit only & Collateral/effect & 0.249 & 0.265 & -0.016 & 0.801 \\
Direct-logit only & Effect $L_2$ & 14.192 & 12.860 & +1.332 & 0.050 \\
Direct-logit only & KL/effect & 0.000053 & 0.000053 & -0.000001 & 0.727 \\
\bottomrule
\end{tabular}%
}
\end{center}
\end{table}

\begin{table}[!htbp]
\caption{Residualized-target prediction results for the controlled Llama Scope 128K setting. Stability targets are residualized against effect magnitude before cross-validated regression. Rows report five-fold CV Spearman correlation and CV $R^2$. The full no-magnitude predictor set remains predictive after residualization, indicating that the Llama 128K stability signal is not explained only by effect size.}
\label{tab:llama128k-residualized}
\begin{center}
\small
\setlength{\tabcolsep}{4pt}
\begin{tabular}{@{}llrrr@{}}
\toprule
Residualized target & Predictor set & $n_{\mathrm{pred}}$ & CV Spearman & CV $R^2$ \\
\midrule
Signed stability & Frequency only & 1 & -0.076 & -0.005 \\
Signed stability & Activation magnitude & 4 & -0.064 & -0.011 \\
Signed stability & Geometry only & 5 & 0.105 & -0.006 \\
Signed stability & Direct-logit only & 5 & 0.107 & 0.028 \\
Signed stability & Coactivation only & 3 & 0.279 & 0.052 \\
Signed stability & Full no-magnitude & 16 & 0.402 & 0.110 \\
Signed stability & Full all & 20 & 0.410 & 0.114 \\
\midrule
Absolute stability & Frequency only & 1 & -0.110 & -0.007 \\
Absolute stability & Activation magnitude & 4 & -0.128 & -0.021 \\
Absolute stability & Geometry only & 5 & 0.106 & 0.005 \\
Absolute stability & Direct-logit only & 5 & 0.161 & 0.043 \\
Absolute stability & Coactivation only & 3 & 0.271 & 0.038 \\
Absolute stability & Full no-magnitude & 16 & 0.361 & 0.092 \\
Absolute stability & Full all & 20 & 0.368 & 0.088 \\
\bottomrule
\end{tabular}
\end{center}
\end{table}

\begin{table}[!htbp]
\caption{Top univariate predictor--target correlations for the controlled Llama Scope 128K setting. Rows show the strongest primary-target relationships by absolute Spearman correlation over 300 sampled SAE features. The strongest relationships involve decoder norm and direct-logit footprint predicting effect magnitude, while collateral-related targets are also associated with activation and direct-logit statistics.}
\label{tab:llama128k-univariate-top}
\begin{center}
\small
\setlength{\tabcolsep}{3pt}
\begin{tabular}{@{}llrr@{}}
\toprule
Predictor & Target & Spearman $\rho$ & Pearson $r$ \\
\midrule
Decoder norm & Effect $L_2$ & 0.638 & 0.684 \\
Direct-logit $L_2$ & Effect $L_2$ & 0.588 & 0.656 \\
Direct-logit $L_\infty$ & Effect $L_2$ & 0.487 & 0.369 \\
Decoder norm & Effect CV & 0.426 & 0.403 \\
Token activation mean & Downstream moved mean & 0.390 & 0.201 \\
Direct-logit $L_2$ & Downstream count & 0.389 & 0.544 \\
Direct-logit $L_2$ & Effect CV & 0.376 & 0.372 \\
Token activation kurtosis & Downstream moved mean & -0.376 & -0.340 \\
Coactivation count mean & Effect $L_2$ & -0.374 & -0.377 \\
Activation entropy & Downstream moved mean & 0.372 & 0.357 \\
Feature frequency & Downstream moved mean & 0.370 & 0.211 \\
Token binary entropy & Downstream moved mean & 0.370 & 0.271 \\
Token activation std. & Downstream moved mean & 0.359 & 0.275 \\
Direct-logit $L_\infty$ & Effect CV & 0.354 & 0.237 \\
Decoder norm & Downstream count & 0.346 & 0.500 \\
\bottomrule
\end{tabular}
\end{center}
\end{table}

%% file: appendix_reproducibility_hardware.tex
\section{Reproducibility and Hardware}
\label{app}

We release anonymized code for reproducing the experiments at:
\begin{center}
\url{https://anonymous.4open.science/r/sae-steering-side-effects-96B5/README.md}
\end{center}
The repository contains the experiment scripts, model/SAE configuration files, figure-generation code, dependency files, and reproduction instructions. Generated result CSV/JSON files and rendered figures are not tracked in the repository; the manuscript and appendices report the final tables and figures, and users can regenerate these artifacts by running the provided scripts.

All experiments were run on NVIDIA A100 GPUs. The experiments use pretrained language models and pretrained sparse autoencoders only; no model fine-tuning or SAE training is performed. The main computational cost comes from forward passes used to cache activations, compute intervention-free predictors, construct steering labels, and evaluate held-out screening groups. The predictive-evaluation stages use Wikitext-103 contexts and fixed random seeds for feature sampling, context construction, and cross-validation splits. The held-out screening stage uses fresh Wikitext validation contexts not used for steering-label construction.